\documentclass[10pt,twocolumn,letterpaper]{article}

\usepackage{iccv}
\usepackage{times}
\usepackage{epsfig}
\usepackage{graphicx}
\usepackage{amsmath}
\usepackage{amssymb}
\usepackage{adjustbox}
\usepackage{multirow}

\usepackage[breaklinks=true,bookmarks=false]{hyperref}

\iccvfinalcopy %

\def\iccvPaperID{****} %

\ificcvfinal\fi

\begin{document}

	\title{What Would You Expect? Anticipating Egocentric Actions With Rolling-Unrolling LSTMs and Modality Attention}
	
	\author{Antonino Furnari \hspace{4mm} Giovanni Maria Farinella \\
		University of Catania - Department of Mathematics and Computer Science\\
		{\tt\small http://iplab.dmi.unict.it/fpv/ -  \{furnari,gfarinella\}@dmi.unict.it}
	}
	
	\maketitle
	\ificcvfinal\thispagestyle{empty}\fi

	\begin{abstract}
		\vspace{-\topsep}
		Egocentric action anticipation consists in understanding which objects the camera wearer will interact with in the near future and which actions they will perform. We tackle the problem proposing an architecture able to anticipate actions at multiple temporal scales using two LSTMs to 1) summarize the past, and 2) formulate predictions about the future. The input video is processed considering three complimentary modalities: appearance (RGB), motion (optical flow) and objects (object-based features). Modality-specific predictions are fused using a novel Modality ATTention (MATT) mechanism which learns to weigh modalities in an adaptive fashion. Extensive evaluations on two large-scale benchmark datasets show that our method outperforms prior art by up to $+7\%$ on the challenging EPIC-Kitchens dataset including more than $2500$ actions, and generalizes to EGTEA Gaze+. Our approach is also shown to generalize to the tasks of early action recognition and action recognition. Our method is ranked first in the public leaderboard of the EPIC-Kitchens egocentric action anticipation challenge 2019. Please see the project web page for code and additional details: \textit{http://iplab.dmi.unict.it/rulstm}.
	\end{abstract}
	\vspace{-2\topsep}

	\section{Introduction}
	Anticipating the near future is a natural task for humans and a fundamental one for intelligent systems when it is necessary to react before an action is completed (e.g., to anticipate a pedestrian crossing the street from an autonomous vehicle~\cite{de2016online}) or even before it starts (e.g., to notify a user who is performing the wrong action in a known workflow~\cite{soran2015generating}).
	Additionally, tasks such as action anticipation~\cite{gao2017red,koppula2016anticipating,vondrick2016anticipating} and early action recognition~\cite{aliakbarian2017encouraging,de2016online,ma2016learning} pose a series of key challenges from a computational perspective. Indeed, methods addressing these tasks need to model the relationships between past, future events and incomplete observations. First Person (Egocentric) Vision~\cite{kanade2012first}, in particular, offers an interesting scenario to study anticipation problems. On one hand, whilst being a natural task for humans, anticipating the future from egocentric video is computationally challenging due to the ability of wearable cameras to acquire long videos of complex activities involving many objects and actions performed by a user from their unique point of view. On the other hand, investigating these tasks is fundamental for the construction of intelligent wearable systems able to anticipate the user's goal and assist them~\cite{kanade2012first}.

	In this paper, we address the problem of egocentric action anticipation.
	\begin{figure}
		\includegraphics[width=\linewidth]{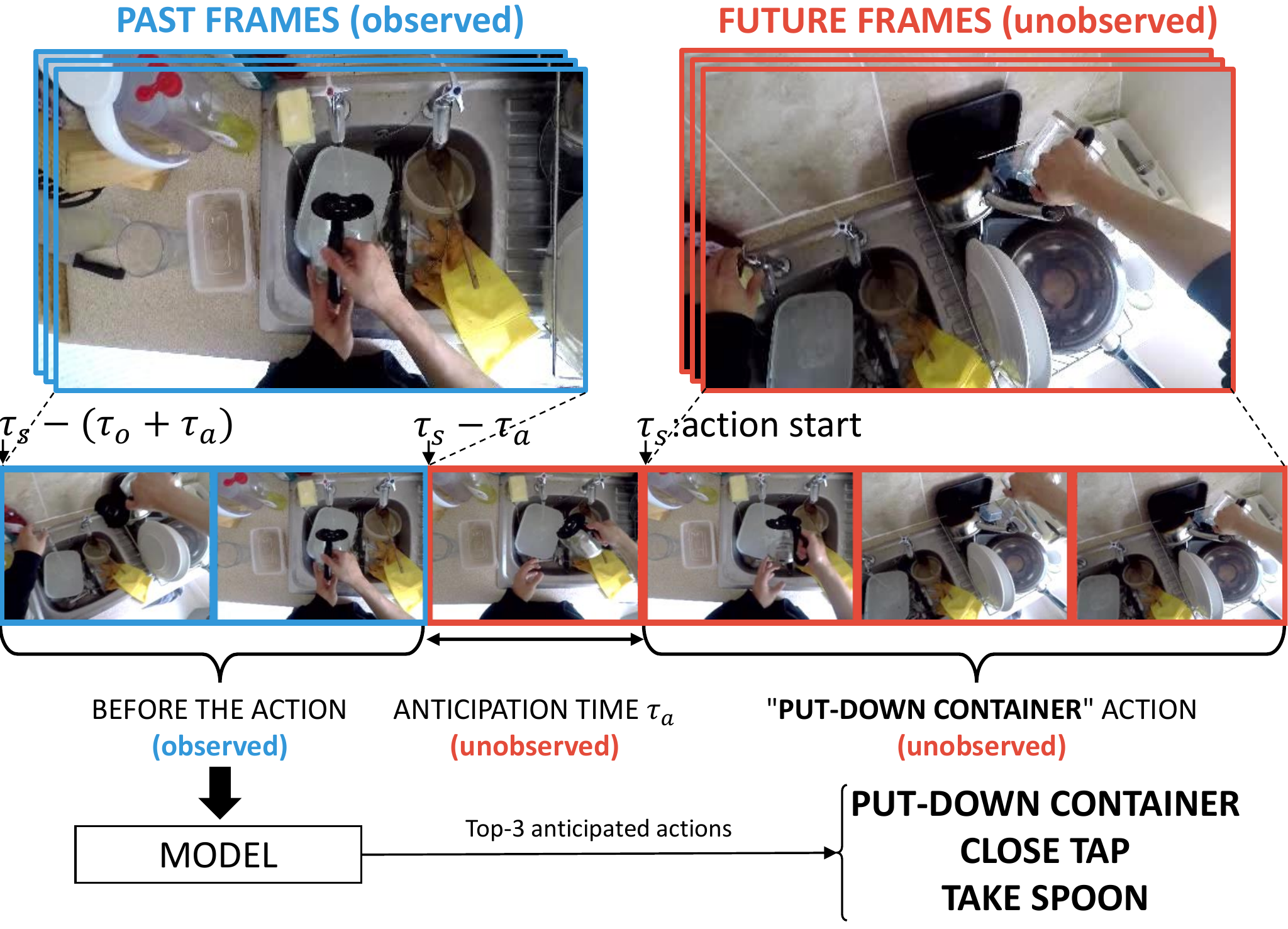}
		\caption{Egocentric Action Anticipation. See text for notation.}
		\label{fig:anticipation}
		\vspace{-\topsep}
	\end{figure}
	As defined in~\cite{Damen2018EPICKITCHENS} and illustrated in \figurename~\ref{fig:anticipation}, the task consists in recognizing an action starting at time $\tau_s$ by observing a video segment \textit{preceding the action} starting at time $\tau_s-(\tau_o+\tau_a)$ and ending at time $\tau_s-\tau_a$, where the ``observation time'' $\tau_o$ indicates the length of the observed segment, whereas the ``anticipation time'' $\tau_a$ denotes how many seconds in advance actions are to be anticipated. 
	While action anticipation has been investigated in classic third person vision~\cite{abu2018will,gao2017red,huang2014action,koppula2016anticipating,vondrick2016anticipating,jain2016recurrent}, less attention has been devoted to the egocentric scenario~\cite{Damen2018EPICKITCHENS,furnari2018Leveraging,Ryoo2015a}. %

	We observe that egocentric action anticipation methods need to address two sub-tasks: 1)~summarizing what has been observed in the past (e.g., ``a container has been washed'' in the observed segment in \figurename~\ref{fig:anticipation}), and 2)~making hypotheses about what will happen in the future (e.g., ``put-down container'', ``close tap'', ``take spoon'' in \figurename~\ref{fig:anticipation}). While previous approaches attempted to address these two sub-tasks jointly~\cite{aliakbarian2017encouraging,Damen2018EPICKITCHENS,ma2016learning,vondrick2016anticipating}, our method disentangles them using two separate LSTMs. The ``Rolling'' LSTM (R-LSTM) is responsible for continuously encoding streaming observations to summarize the past. When the method is required to anticipate actions, the ``Unrolling'' LSTM (U-LSTM) takes over the current hidden and cell states of the R-LSTM and makes predictions about the future. 
	Differently from previous works which considered a fixed anticipation time~\cite{Damen2018EPICKITCHENS,furnari2018Leveraging,vondrick2016anticipating}, the proposed method can anticipate an action at multiple anticipation times, e.g., $2s$, $1.5s$, $1s$ and $0.5s$ before it occurs. The network is pre-trained with a novel ``Sequence Completion Pre-training'' (SCP) technique, which encourages the disentanglement of the two sub-tasks. To take advantage of the complementary nature of different input modalities, the proposed Rolling-Unrolling LSTM (RU) processes spatial observations (RGB frames), motion (optical flow), as well as object-based features. 
	Multimodal predictions are fused with a novel ``Modality ATTention'' mechanism (MATT), which adaptively estimates optimal fusion weights for each modality by considering the outputs of the modality-specific R-LSTM components. 
	Experiments on two large-scale datasets of egocentric videos, EPIC-KTICHENS~\cite{Damen2018EPICKITCHENS} and EGTEA Gaze+~\cite{Li_2018_ECCV}, show that the proposed method outperforms several state-of-the-art approaches and baselines in the task of egocentric action anticipation and generalizes to the tasks of early action recognition and action recognition. 

	The contributions of this work are as follows: 1) we are the first to systematically investigate the problem of egocentric action anticipation within the framework of the challenge proposed in~\cite{Damen2018EPICKITCHENS}; 2) our investigation benchmarks popular ideas and approaches to action anticipation and leads to the definition of RU, an architecture able to anticipate egocentric actions at multiple temporal scales; 3) the proposed model is shown to benefit from two techniques specific to the investigated problem, i.e., i) ``Sequence Completion Pre-training'' and ii) adaptive fusion of multi-modal predictions through Modality ATTention; 4) extensive evaluations highlight the limits of previous approaches and show significant improvements of the proposed method over the state of the art. To support future research, the code implementing the proposed method will be released upon publication.

	\section{Related Work}
	\noindent
	\textbf{Action Recognition\hspace{0.5em}} Our work is related to previous research on action recognition from third person vision~\cite{carreira2017quo,feichtenhofer2017spatiotemporal,feichtenhofer2016convolutional,karpathy2014large,laptev2005space,laptev2008learning,simonyan2014two,tran2015learning,wang2013dense,wang2013action,wang2016temporal,zhou2018temporal} and first person vision~\cite{fathi2011understanding,fathi2012learning,Li_2018_ECCV,li2015delving,ma2016going,pirsiavash2012detecting,ryoo2015pooled,singh2017trajectory,spriggs2009temporal,sudhakaran2018lsta,sudhakaran2018attention}. Specifically, we build on previous ideas investigated in the context of action recognition such as the use of multiple modalities for video analysis~\cite{simonyan2014two}, the use of Temporal Segment Networks~\cite{wang2016temporal} as a principled way to train CNNs for action recognition, as well as the explicit encoding of object-based features~\cite{fathi2011understanding,ma2016going,pirsiavash2012detecting,singh2016first,sudhakaran2018attention} to analyze egocentric video. However, in contrast with the aforementioned works, we address the problem of egocentric action \textit{anticipation} and show that approaches designed for action recognition, such as TSN~\cite{wang2016temporal} and early/late fusion to merge multi-modal predictions~\cite{simonyan2014two} are not directly applicable to the problem of egocentric action anticipation. 
	
	\vspace{0.5\topsep}
	\noindent
	\textbf{Early Action Recognition in Third Person Vision\hspace{0.5em}}
	Early action recognition consists in recognizing an ongoing action as early as possible from partial observations~\cite{de2016online}. The problem has been widely investigated in the domain of third person vision~\cite{aliakbarian2017encouraging,becattini2017done,cao2013recognize,de2016online,de2018modeling,hoai2014max,huang2014sequential,ma2016learning,ryoo2011human}. Differently from these works, we address the task of \textit{anticipating} actions from egocentric video, i.e., the action should be recognized before it starts, hence it cannot even be partially observed at the time of prediction. Given the similarity between early recognition and anticipation, we consider and evaluate some ideas investigated in the context of early action recognition, such as the use of LSTMs to process streaming video~\cite{aliakbarian2017encouraging,de2018modeling,ma2016learning}, and the use of dedicated loss functions~\cite{ma2016learning}. Moreover, we show that the proposed architecture also generalizes to the problem of early egocentric action recognition, achieving state-of-the-art performances.
	
	\vspace{0.5\topsep}
	\noindent
	\textbf{Action Anticipation in Third Person Vision\hspace{0.5em}}
	Action anticipation is the task of predicting an action \emph{before} it occurs~\cite{gao2017red}. Previous works investigated different forms of action and activity anticipation from third person video~\cite{abu2018will,felsen2017will,gao2017red,huang2014action,jain2015car,jain2016recurrent,kitani2012activity,koppula2016anticipating,lan2014hierarchical,mahmud2017joint,vondrick2016anticipating,zeng2017visual}. While we consider the problem of action anticipation from egocentric visual data, our work builds on some ideas explored in past works such as the use of LSTMs to anticipate actions~\cite{abu2018will,gao2017red,jain2016recurrent}, the use of the encoder-decoder framework to encode past observations and produce hypotheses of future actions~\cite{gao2017red}, and the use of object specific features~\cite{mahmud2017joint} to determine which objects are present in the scene. Additionally, we show that other approaches, such as the direct regression of future representations~\cite{gao2017red,vondrick2016anticipating}, do not achieve satisfactory performance in the considered scenario.
	
	\vspace{0.5\topsep}
	\noindent
	\textbf{Anticipation in First Person Vision\hspace{0.5em}}
	Past works on anticipation from egocentric visual data have investigated different problems and considered different evaluation frameworks~\cite{bokhari2016long,chenyou2017forecasting,Furnari2017,SooPark2016,rhinehart2017first,Ryoo2015a,singh2016krishnacam,soran2015generating,zhang2017deep,Zhou2015}. Instead, we tackle the egocentric action anticipation challenge recently proposed in~\cite{Damen2018EPICKITCHENS}, which has been little investigated so far~\cite{furnari2018Leveraging}. While a direct comparison of the proposed approach with the aforementioned works is unfeasible due to the lack of a common framework, our method incorporates some ideas from past approaches, such as the analysis of past actions~\cite{Ryoo2015a} and the detection of the objects present in the scene to infer future actions~\cite{Furnari2017}.

	\begin{figure}
		\includegraphics[width=\linewidth]{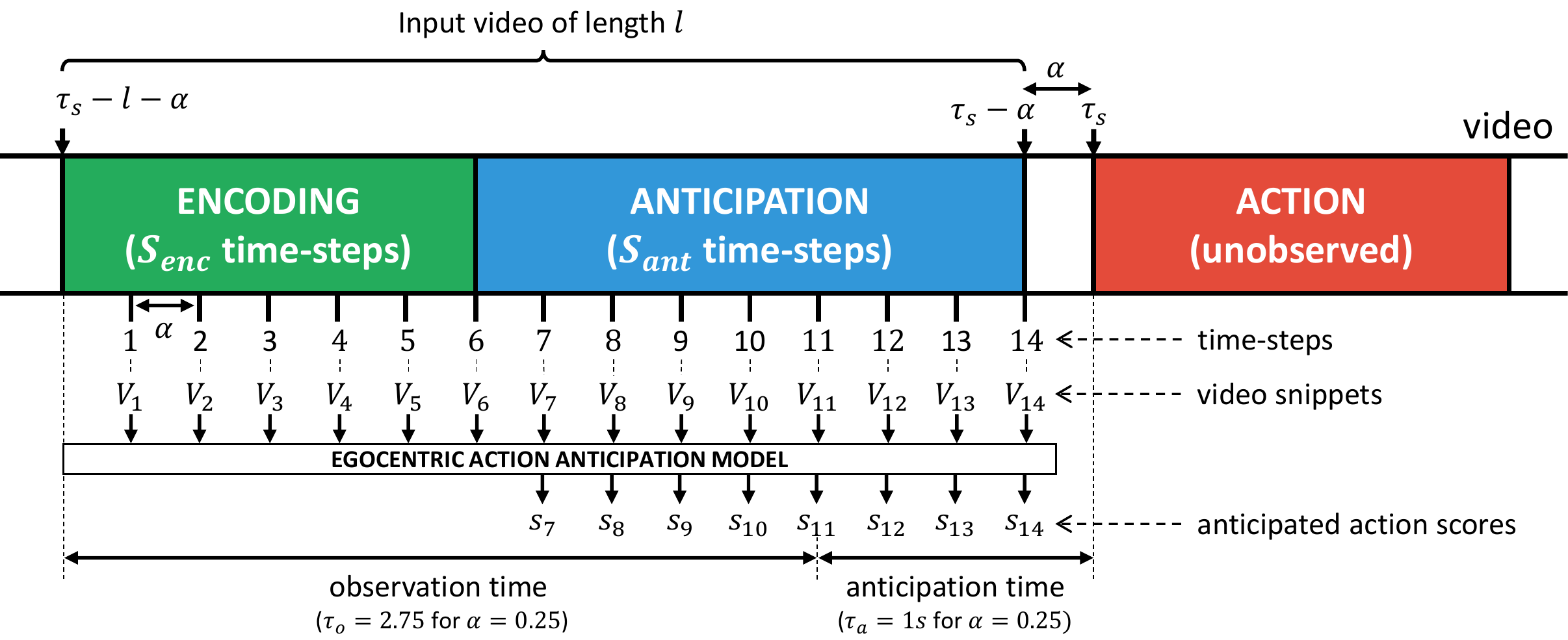}
		\caption{Video processing scheme of the proposed method with $S_{enc}=6$ and $S_{ant}=8$.}
		\label{fig:processing}
		\vspace{-\topsep}
	\end{figure}
	\section{Proposed Approach}
	\label{sec:method}
	
	\noindent
	\textbf{Processing Strategy \hspace{0.5em}}
	\figurename~\ref{fig:processing} illustrates the processing strategy adopted by the proposed method. The video is processed in an on-line fashion, with a short video snippet $V_t$ consumed every $\alpha$ seconds, where $t$ indexes the current time-step. Specifically, an action occurring at time $\tau_s$ is anticipated by processing a video segment of length $l$ starting at time $\tau_s-l-\alpha$ and ending at time $\tau_s-\alpha$. The input video ends at time $\tau_s-\alpha$ as our method aims at anticipating actions \textit{at least} $\alpha$ seconds before they occur. The processing is performed in two stages: an ``encoding'' stage, which is carried out for $S_{enc}$ time-steps, and an ``anticipation'' stage, which is carried out for $S_{ant}$ time-steps. In the encoding stage, the model summarizes the semantic content of the $S_{enc}$ input video snippets without producing any prediction, whereas in the anticipation stage the model continues to encode the semantics of the $S_{ant}$ input video snippets and outputs $S_{ant}$ action scores $s_t$ which can be used to perform action anticipation. This scheme effectively allows to formulate $S_{ant}$ predictions for a single action at multiple anticipation times.
	In our experiments, we set $\alpha=0.25s$, $S_{enc}=6$ and $S_{ant}=8$. In these settings, the model analyzes video segments of length $l=\alpha(S_{enc}+S_{ant})=3.5s$ and outputs $8$ predictions at the following anticipation times: $\tau_a \in \{2s, 1.75s, 1.5s, 1.25s, 1s, 0.75s, 0.5s, 0.25s\}$. It should be noted that, since the predictions are produced \textit{while} processing the video, at time step $t$ the effective observation time will be equal to $\alpha \cdot t$. Hence, the $8$ predictions are performed at the following effective observation times: $\tau_o \in \{1.75s, 2s, 2.25s, 2.5s, 2.75s, 3s, 3.25s, 3.5s\}$. Our formulation generalizes the one proposed in~\cite{Damen2018EPICKITCHENS}, which is illustrated in~\figurename~\ref{fig:anticipation}. For instance, at time-step $t=11$, our model will anticipate actions with an effective observation time equal to $\tau_o=\alpha \cdot t=2.75s$ and an anticipation time equal to $\tau_a=\alpha(S_{ant}+S_{enc}+1-t)=1s$, as in~\cite{Damen2018EPICKITCHENS}.
	
	\vspace{\topsep}
	\noindent
	\textbf{Rolling-Unrolling LSTM\hspace{0.5em}}
	The proposed method uses two separate LSTMs to encode past observations and formulate predictions about the future. Following previous literature~\cite{simonyan2014two}, we include multiple identical branches which analyze the video according to different modalities. Specifically, at each time-step $t$, the input video snippet $V_t$ is represented using different modality-specific representation functions $\varphi_1,\ldots,\varphi_M$ depending on learnable parameters $\theta^{\varphi_1},\ldots,\theta^{\varphi_M}$. This process allows to obtain the modality-specific feature vectors $f_{1,t}=\varphi_1(V_t),\ldots,f_{M,t}=\varphi_M(V_t)$, where $M$ is the total number of modalities (i.e., the total number of branches in our architecture), and $f_{m,t}$ is the feature vector computed at time-step $t$ for the modality $m$. The feature vector $f_{m,t}$ is fed to the $m^{th}$ branch of the architecture. While our model can easily incorporate different modalities, in this work we consider $M=3$ modalities, i.e., RGB frames (spatial branch), optical flow (motion branch) and object-based features (object branch).

	\begin{figure}
		\includegraphics[width=\linewidth]{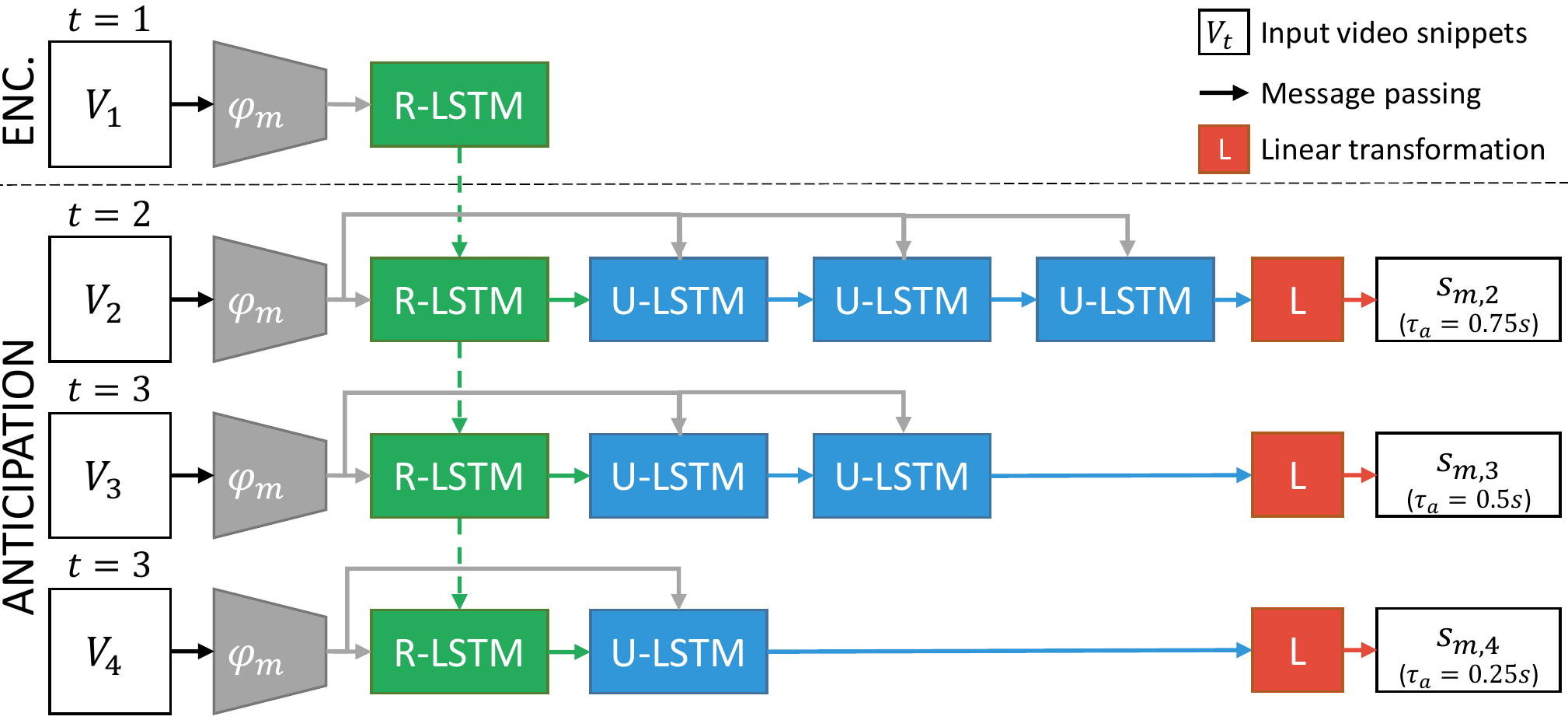}
		\caption{Example of RU modality-specific branch with $S_{enc}=1$ and $S_{ant}=3$.}
		\label{fig:branch}
		\vspace{-\topsep}
	\end{figure}

	\figurename~\ref{fig:branch} illustrates the processing taking place in a single branch $m$ of the proposed RU model. For illustration purposes only, the figure shows an example in which $S_{enc}=1$ and $S_{ant}=3$. At time step $t$, the feature vector $f_{m,t}$ is fed to the Rolling LSTM (R-LSTM), which encodes its semantic content recursively, as follows:
	\begin{equation}\small
	(h_{m,t}^R,c_{m,t}^R)=LSTM_{\theta_m^R}(f_{m,t},h_{m,t-1}^R,c_{m,t-1}^R)
	\end{equation}
	where $LSTM_{\theta_m^R}$ denotes the R-LSTM of branch $m$, depending on the learnable parameters $\theta_m^R$, whereas $h^R_{m,t}$ and $c^R_{m,t}$ are the hidden and cell states computed at time $t$ in the modality $m$. The initial hidden and cell states of the R-LSTM are initialized with zeros: $h^R_{m,0}=\textbf{0},\ c^R_{m,0}=\textbf{0}$.

	In the anticipation stage, at time step $t$, the Unrolling LSTM (U-LSTM) is used to make predictions about the future. The U-LSTM takes over the hidden and cell vectors of the R-LSTM at the current time-step (i.e., $h_{m,t}^R$ and $c_{m,t}^R$) and iterates over the representation of the current video snippet $f_{m,t}$ for a number of times $n_t$ equal to the number of time-steps required to reach the beginning of the action, i.e., $n_t=S_{ant}+S_{enc}-t+1$. Hidden and cell states of the U-LSTM are computed as follows at the $j^{th}$ iteration:
	\begin{equation}\small
	(h_{m,j}^U,c_{m,j}^U)=LSTM_{\theta_m^U}(f_{m,t},h_{m,j-1}^U,c_{m,j-1}^U)
	\label{eq:ulstm}
	\end{equation}
	where $LSTM_{\theta_m^U}$ is the U-LSTM of branch $m$, depending on the learnable parameters $\theta_m^U$, and $h^U_{m,t}$, $c^U_{m,t}$ are the hidden and cell states computed at iteration $j$ for the modality $m$. The initial hidden and cell states of the U-LSTM are initialized from the current hidden and cell states computed by the R-LSTM: $h^U_{m,0}=h^R_{m,t},\ c^U_{m,0}=c^R_{m,t}$. Note that the input $f_{m,t}$ of the U-LSTM does not depend on $j$ (see eq.~\eqref{eq:ulstm}) because it is fixed during the ``unrolling'' procedure. The main rationale of ``unrolling'' the U-LSTM for a different number of times at each time-step is to encourage it to differentiate predictions at different anticipation times.
	
	Modality-specific action scores $s_{m,t}$ are computed at time-step $t$ by processing the last hidden vector of the U-LSTM with a linear transformation with learnable parameters $\theta^W_m$ and $\theta^b_m$: $s_{m,t}=\theta^W_m h^U_{m,n_t}+\theta^b_m.$

	\begin{figure}
		\includegraphics[width=\linewidth]{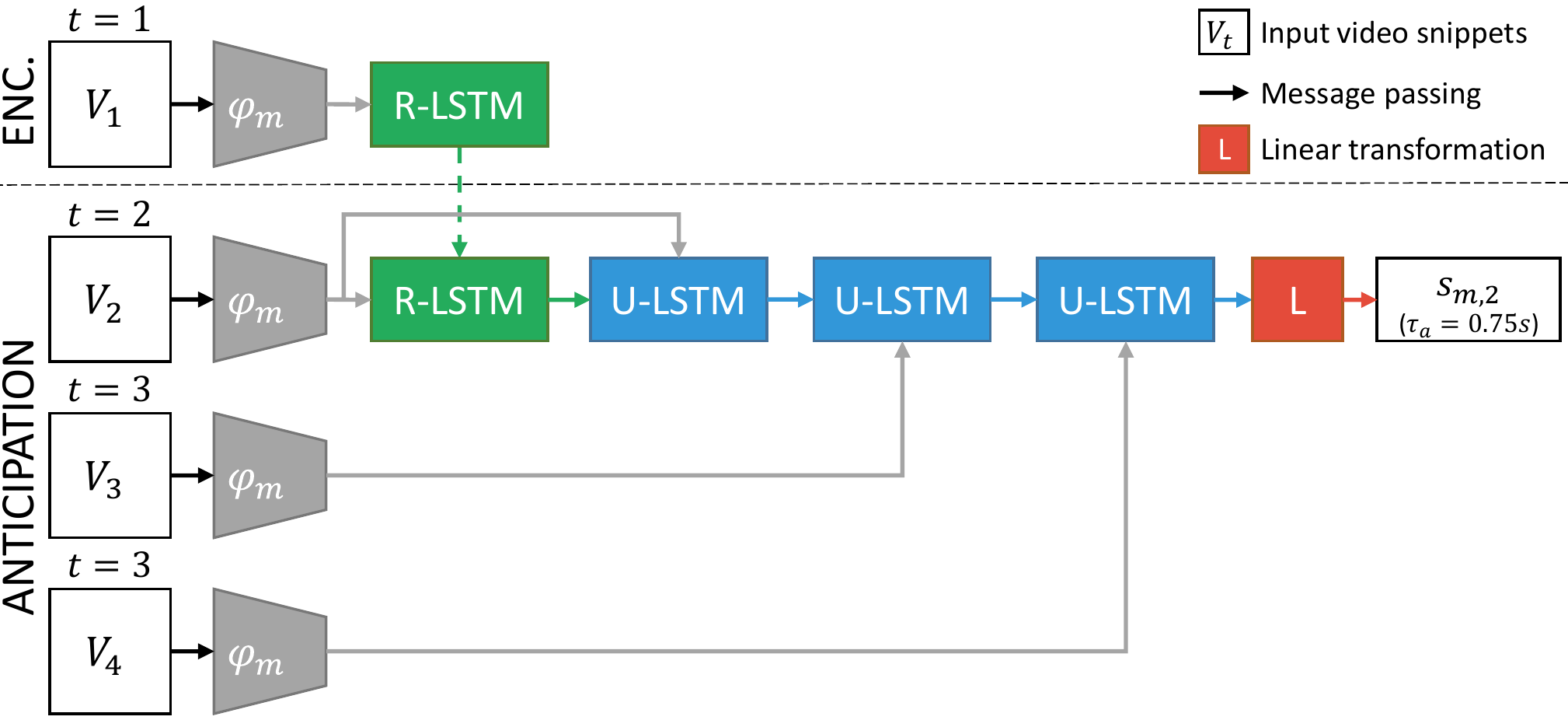}
		\caption{Example of connection scheme used during SCP for time-step $t=2$.}
		\label{fig:sequence_completion}
		\vspace{-\topsep}
	\end{figure}
	
	\vspace{\topsep}
	\noindent
	\textbf{Sequence Completion Pre-Training (SCP)\hspace{0.5em}}
	The two LSTMs composing the RU architecture are designed to address two specific sub-tasks: the R-LSTM is responsible for encoding past observations and summarizing what has happened up to a given time-step, whereas the U-LSTM focuses on anticipating future actions conditioned on the hidden and cell vectors of the R-LSTM. To encourage the two LSTMs to specialize on the two different sub-tasks, we propose to train the architecture using a novel Sequence Completion Pre-training (SCP) procedure. During SCP, the connections of the U-LSTM are modified to allow it to process future representations, rather than iterating on the current one. In practice, the U-LSTM hidden and cell states are computed as follows during SCP:
	\begin{equation}\small
	(h_{m,j}^U,c_{m,j}^U)=LSTM_{\theta_m^U}(f_{m,t+j-1},h_{m,j-1}^U,c_{m,j-1}^U)
	\end{equation}
	where the input representations $f_{m,t+j-1}$ are sampled from future time-steps $t+j-1$. \figurename~\ref{fig:sequence_completion} illustrates an example of the connection scheme used during SCP for time-step $t=2$.
	The main goal of pre-training the RU with SCP is to allow the R-LSTM to focus on summarizing past representations without trying to anticipate the future. 

	\begin{figure}
		\includegraphics[width=\linewidth]{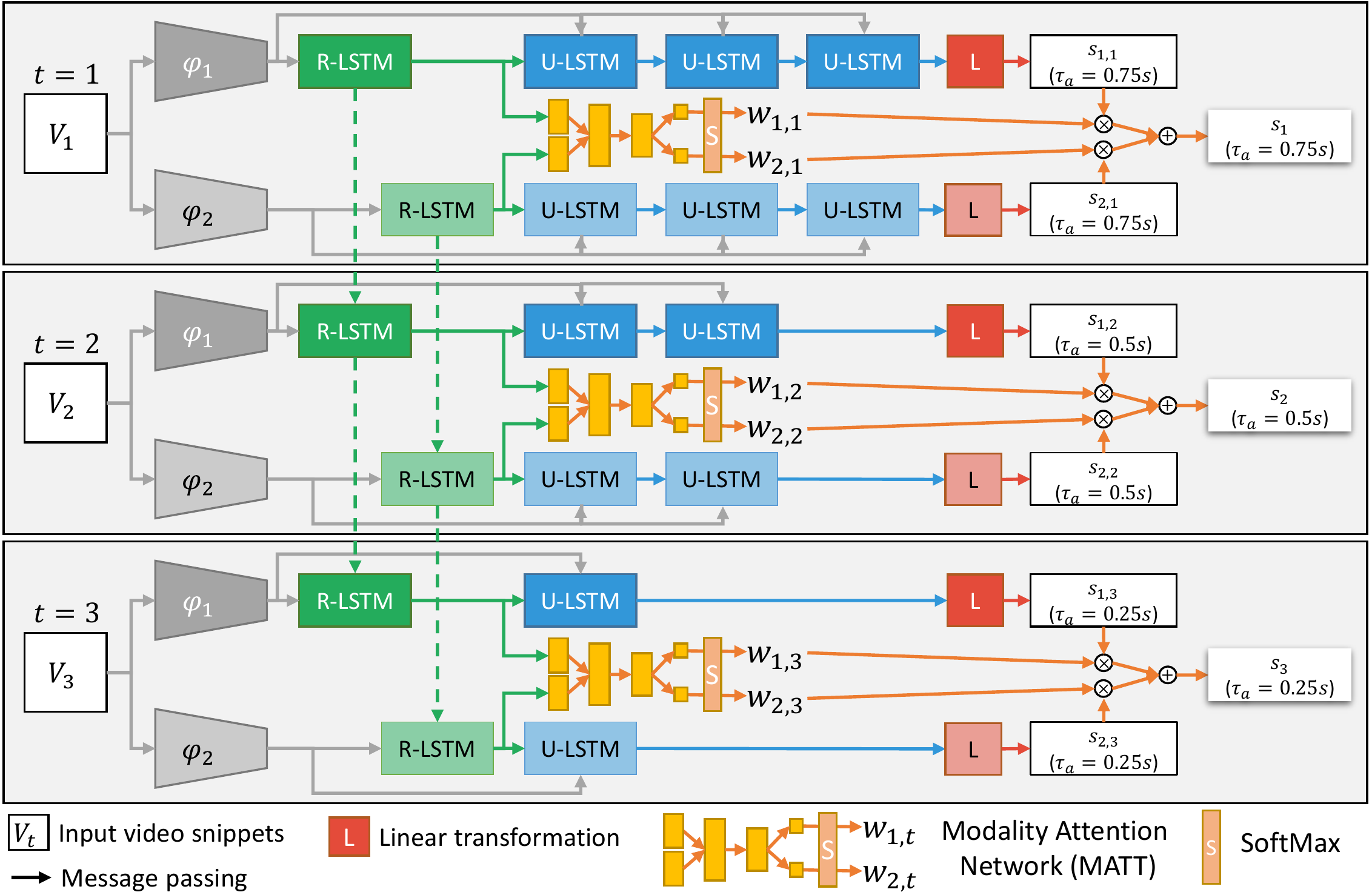}
		\caption{Example of the complete RU architecture with two modalities and the Modality ATTention mechanism (MATT).}
		\label{fig:matt}
		\vspace{-\topsep}
	\end{figure}
	
	\vspace{\topsep}
	\noindent
	\textbf{Modality ATTention (MATT)\hspace{0.5em}}
	Coherently with past work on egocentric action anticipation~\cite{Damen2018EPICKITCHENS}, we found it sub-optimal to fuse multi-modal predictions with classic approaches such as early and late fusion. This is probably due to the fact that, when anticipating egocentric actions, one modality might be more useful than another (e.g., appearance over motion), depending on the processed sample. Inspired by previous work on attention~\cite{bahdanau2014neural,xu2015show} and multi-modal fusion~\cite{mees16iros}, we introduce a Modality ATTention (MATT) module which computes a set of attention scores indicating the relative importance of each modality for the final prediction. At a given time-step $t$, such scores are obtained by processing the concatenation of the hidden and cell vectors of the R-LSTM networks belonging to all branches $m=1,\ldots,M$ with a deep neural network $D$ depending on the learnable parameters $\theta^{MATT}$:
	\begin{equation}\small
	\lambda_t=D_{\theta^{MATT}}(\oplus_{m=1}^M(h_{m,t}^R \oplus c_{m,t}^R))
	\end{equation}
	where $\oplus$ denotes the concatenation operator and $\oplus_{m=1}^M(h_{m,t}^R \oplus c_{m,t}^U)$ is the concatenation of the hidden and cell vectors produced by the R-LSTM at time-step $t$ across all modalities. Late fusion weights can be obtained normalizing the score vector $\lambda_t$ with the softmax function in order to make sure that fusion weights sum to one: $
	w_{m,t}=\frac{exp(\lambda_{t,m})}{\sum_k exp(\lambda_{t,k})}
	$,
	where $\lambda_{t,m}$ is the $m^{th}$ component of the score vector $\lambda_t$. The final set of fusion weights is obtained at time-step $t$ by merging the modality-specific predictions produced by the different branches with a linear combination as follows: $s_t=\sum_m w_{m,t} \cdot s_{m,t}$.
	\figurename~\ref{fig:matt} illustrates an example of a complete RU with two modalities and the MATT fusion mechanism. For illustration purposes, the figure shows only three anticipation steps.

	\vspace{\topsep}
	\noindent
	\textbf{Branches and Representation Functions\hspace{0.5em}}
	We instantiate the proposed architecture with $3$ branches: a spatial branch which processes RGB frames, a motion branch which processes optical flow, and an object branch which processes object-based features. Our architecture analyzes video snippets of $5$ frames $V_t=\{I_{t,1},I_{t,2},\ldots,I_{t,5}\}$, where $I_{t,i}$ is the $i^{th}$ frame of the video snippet. The representation function $\varphi_1$ of the spatial branch computes the feature vector $f_{1,t}$ by extracting features from the last frame $I_{t,5}$ of the video snippet using a Batch Normalized Inception CNN~\cite{ioffe2015batch} trained for action recognition. The representation function $\varphi_2$ of the motion branch extracts optical flow from the $5$ frames of the current video snippet as proposed in~\cite{wang2016temporal}. The computed $x$ and $y$ optical flow form a tensor with $10$ channels, which is fed to a Batch Normalized Inception CNN trained for action recognition to obtain the feature vector $f_{2,t}$. Note that $\varphi_1$ and $\varphi_2$ allow to obtain ``action-centric'' representations of the input frame which can be used by the R-LSTM to summarize what has happened in the past. The representation function $\varphi_3$ of the object branch extracts objects from the last frame $I_{t,5}$ of the input snippet $V_t$ using an object detector. A fixed-length representation $f_{3,t}$ is obtained by accumulating the confidence scores of all bounding boxes predicted for each object class. Specifically, let $b_{t,i}$ be the $i^{th}$ bounding box detected in image $I_{t,5}$, let $b^c_{t,i}$ be its class and let $b^s_{t,i}$ be its detection confidence score. The $j^{th}$ component of the output representation vector $f_{3,t}$ is obtained by summing the confidence scores of all detected objects of class $j$, i.e., $f_{3,t,j}=\sum_i [b^c_{t,i}=j]b^s_{t,i}$, where $[\cdot]$ denotes the Iverson bracket. This representation only encodes the presence of an object in the scene, discarding its position in the frame, similarly to the representation proposed in~\cite{pirsiavash2012detecting} for egocentric activity recognition. We found this holistic representation to be sufficient in the case of egocentric action anticipation. Differently from $\varphi_1$ and $\varphi_2$, $\varphi_3$ produces object-centric features, which carry information on what objects are present in the scene and hence could be interacted next.\footnote{See supp. for details on implementation and training of our method.}

	\vspace{-\topsep}
	\section{Experimental Settings}
	\noindent
	\textbf{Datasets\hspace{0.5em}}
	We perform experiments on two large-scale datasets of egocentric videos: EPIC-Kitchens~\cite{Damen2018EPICKITCHENS} and EGTEA Gaze+~\cite{Li_2018_ECCV}. EPIC-Kitchens contains $39,596$ action annotations, $125$ verbs, and $352$ nouns. We split the public training set of EPIC-Kitchens ($28,472$ action segments) into training ($23,493$ segments) and validation ($4,979$ segments) sets by randomly choosing $232$ videos for training and $40$ videos for validation. We considered all unique $(verb, noun)$ pairs in the public training set, thus obtaining $2,513$ unique actions. EGTEA Gaze+ contains $10,325$ action annotations, $19$ verbs, $51$ nouns and $106$ unique actions. Methods are evaluated on EGTEA Gaze+ reporting the average performance across the three splits provided by the authors of the dataset~\cite{Li_2018_ECCV}.
	
	\vspace{\topsep}
	\noindent
	\textbf{Evaluation Measures\hspace{0.5em}}
	Methods are evaluated using Top-k evaluation measures, i.e., we deem a prediction correct if the ground truth action falls in the top-k predictions. As observed in previous works~\cite{furnari2018Leveraging,koppula2016anticipating}, this evaluation scheme is appropriate given the uncertainty of future predictions (i.e., many plausible actions can follow an observation).
	Specifically, we use the Top-5 accuracy as a class-agnostic measure and Mean Top-5 Recall as a class aware metric. Top-5 recall~\cite{furnari2018Leveraging} for a given class $c$ is defined as the fraction of samples of ground truth class $c$ for which the class $c$ is in the list of the top-5 anticipated actions. Mean Top-5 Recall averages Top-5 recall values over classes. When evaluating on EPIC-Kitchens, Top-5 Recall is averaged over the provided list of many-shot verbs, nouns and actions. Results on the EPIC-Kitchens official test set are reported using the suggested evaluation measures, i.e., Top-1 accuracy, Top-5 accuracy, Precision and Recall.
	
	To assess the timeliness of anticipations, we design a new evaluation measure inspired by the AMOC curve~\cite{hoai2014max}. Let $s_t$ be the score anticipated at time-step $t$ for an action of ground truth class $c$, let $\tau_t$ be the anticipation time at time-step $t$, and let $tk(s_t)$ be the set of top-$k$ actions as ranked by the scores $s_t$. We define as ``time to action'' at rank $k$ the largest anticipation time (i.e., the time of earliest anticipation) in which a correct prediction has been made according to the top-$k$ criterion: $TtA(k)=max\{\tau(s_t) |c \in tk(s_t),\ \forall t\}$.
	If an action is not correctly anticipated in any of the time-steps, we set $TtA(k)=0$. 
	The mean time to action over the whole test set $mTtA(k)$ indicates how early, in average, a method can anticipate actions.
	
	Performances are evaluated for verb, noun and action predictions on the EPIC-Kitchens dataset. We obtain verb and noun scores by marginalization over action scores for all methods except the one proposed in~\cite{Damen2018EPICKITCHENS}, which predicts verb and noun scores directly.
	
	\vspace{\topsep}
	\noindent
	\textbf{Compared Methods\hspace{0.5em}}
	We compare the proposed method with respect to $7$ state-of-the approaches and baselines. Specifically, we consider the Deep Multimodal Regressor (DMR) proposed in~\cite{vondrick2016anticipating}, the Anticipation Temporal Segment Network (ATSN) of~\cite{Damen2018EPICKITCHENS}, the anticipation Temporal Segment Network trained with verb-noun Marginal Cross Entropy Loss (MCE) described in~\cite{furnari2018Leveraging}, and the Encoder-Decoder LSTM (ED) introduced in~\cite{gao2017red}. We also consider baselines obtained adapting approaches proposed for early action recognition to the problem of egocentric action anticipation: the Feedback Network LSTM (FN) proposed in~\cite{de2018modeling}, a single LSTM architecture (we use the same parameters of our R-LSTM) trained using the Ranking Loss on Detection Score proposed in~\cite{ma2016learning} (RL), and an LSTM trained using the Exponential Anticipation Loss proposed in~\cite{jain2016recurrent} (EL). These baselines adopt the video processing scheme illustrated in \figurename~\ref{fig:processing} and are implemented as two stream networks with a spatial and a temporal branch whose predictions are fused by late fusion.\footnote{See supp. for implementation details of the considered methods.}

	\begin{table*}
		\begin{adjustbox}{width=\linewidth,center}
			\begin{tabular}{p{1.2cm}cccccccc||ccc|ccc|ccc}
				\hline
				\multicolumn{1}{c}{} & \multicolumn{8}{c||}{Top-5 ACTION Accuracy\% @ different $\tau_a$(s)} & \multicolumn{3}{c|}{Top-5 Acc.\% @1s} & \multicolumn{3}{c|}{M. Top-5 Rec.\% @1s} & \multicolumn{3}{c}{Mean $TtA(5)$}   \\  \hline %
				\multicolumn{1}{c}{} & $2$ & $1.75$ & $1.5$ & $1.25$ & $1.0$& $0.75$&$0.5$&$0.25$ & VERB & NOUN & ACT. & VERB & NOUN & ACT. & VERB & NOUN & ACT.\\ \hline
				DMR~\cite{vondrick2016anticipating} & /              & /              & /              & /              & 16.86          & /              & /              & /              & 73.66          & 29.99          & 16.86          & 24.50          & 20.89          & 03.23          & /              & /              & /              \\
				ATSN~\cite{Damen2018EPICKITCHENS}  & /              & /              & /              & /              & 16.29          & /              & /              & /              & \underline{77.30} & 39.93          & 16.29          & 33.08          & 32.77          & 07.60          & /              & /              & /              \\
				MCE~\cite{furnari2018Leveraging}  & /              & /              & /              & /              & 26.11          & /              & /              & /              & 73.35          & 38.86          & 26.11          & 34.62          & 32.59          & 06.50          & /              & /              & /              \\
				ED~\cite{gao2017red}           & 21.53          & 22.22          & 23.20          & 24.78          & 25.75          & 26.69          & 27.66          & 29.74          & 75.46          & 42.96          & 25.75          & \underline{41.77}          & \underline{42.59}          & \underline{10.97}          & \underline{01.60} & \underline{01.02} & \underline{00.63} \\
				FN~\cite{de2018modeling}       & 23.47          & 24.07          & 24.68          & 25.66          & 26.27          & 26.87          & 27.88          & 28.96          & 74.84          & 40.87          & 26.27          & 35.30          & 37.77          & 06.64          & 01.52          & 00.86          & 00.56          \\
				RL~\cite{ma2016learning}       & \underline{25.95} & \underline{26.49} & \underline{27.15} & \underline{28.48} & \underline{29.61} & \underline{30.81} & \underline{31.86} & 32.84          & 76.79          & \underline{44.53} & \underline{29.61} & 40.80 & 40.87 & 10.64 & 01.57          & 00.94          & 00.62          \\
				EL~\cite{jain2016recurrent}    & 24.68          & 25.68          & 26.41          & 27.35          & 28.56          & 30.27          & 31.50          & \underline{33.55} & 75.66          & 43.72          & 28.56          & 38.70          & 40.32          & 08.62          & 01.55          & 00.94          & 00.62          \\
				\textbf{RU}                             & \textbf{29.44} & \textbf{30.73} & \textbf{32.24} & \textbf{33.41} & \textbf{35.32} & \textbf{36.34} & \textbf{37.37} & \textbf{38.98} & \textbf{79.55} & \textbf{51.79} & \textbf{35.32} & \textbf{43.72} & \textbf{49.90} & \textbf{15.10} & \textbf{01.62} & \textbf{01.11} & \textbf{00.76} \\ \hline
				Improv.                                 & +3.49          & +4.24          & +5.09          & +4.93          & +5.71          & +5.53          & +5.51          & +5.43          & +2.25          & +7.26          & +5.71          & +1.95          & +7.31          & +4.13          & +0.02          & +0.09          & +0.13          \\ \hline
			\end{tabular}
		\end{adjustbox}
		\caption{Egocentric action anticipation results on the EPIC-KITCHENS dataset}
		\label{tab:anticipation_ek}
		\vspace{-\topsep}
	\end{table*}

	\section{Results}
	\vspace{-0.5\topsep}
	\noindent
	\textbf{Anticipation Results on EPIC-KITCHES\hspace{0.5em}} \tablename~\ref{tab:anticipation_ek} compares RU with respect to the state-of-the-art on our EPIC-Kitchens validation set. The left part of the table reports Top-5 action accuracy for the $8$ anticipation times. Note that some methods~\cite{Damen2018EPICKITCHENS,furnari2018Leveraging,vondrick2016anticipating} can anticipate actions only at a fixed anticipation time. The right part of the table breaks down Top-5 accuracy and Mean Top-5 Recall for verbs, nouns and actions, for anticipation time $\tau_a=1s$, and reports mean $TtA(5)$ scores. Best results are in bold, whereas second-best results are underlined. The last row reports the improvement of RU with respect to second best results.

	The proposed approach outperforms all competitors by consistent margins according to all evaluation measures, reporting an average improvement over prior art of about $5\%$ on Top-5 action accuracy with respect to all anticipation times. Note that this margin is significant given the large number of $2,513$ action classes present in the dataset. 
	Methods based on TSN (ATSN and MCE) generally achieve low performance, which suggests that simply adapting action recognition methods to the problem of anticipation is insufficient. 
	Interestingly, DMR and ED, which are explicitly trained to anticipate future representations, achieve sub-optimal Top-5 action accuracy as compared to methods trained to predict future actions directly from input images (e.g., compare DMR with MCE, and ED with FN/RL/EL). This might be due to the fact that anticipating future representations is particularly challenging in the case of egocentric video, where the visual content changes continuously because of the mobility of the camera. 
	RL consistently achieves second best results with respect to all anticipation times, except $\tau_a=0.25$, where it is outperformed by EL. 
	RU is particularly strong on nouns, achieving a Top-5 noun accuracy of $51.79\%$ and a mean Top-5 noun recall of $49.90\%$, which improves over prior art by $+7.26\%$ and $+7.31\%$. The small drop in performance when passing from class-agnostic measures to class-aware measures (i.e., $51.79\%$ to $49.90\%$) suggests that our method does not over-fit to the distribution of nouns of the training set. It is worth noting that mean Top-5 Recall values are averaged over fairly large sets of $26$ many-shot verbs, $71$ many-shot nouns, and $819$ many-shot actions, as specified in~\cite{Damen2018EPICKITCHENS}. Differently, all compared methods exhibit large drops in verb and action performance when passing from class-agnostic to class-aware measures. Our insight into this different pattern is that anticipating the next object which will be used (i.e., anticipating nouns) is much less ambiguous than anticipating the way in which the object will be used (i.e., anticipating verbs and actions). It is worth noting that second best Top-5 verb and noun accuracy are achieved by different methods (i.e., ATSN in the case of verbs and RL in the case of nouns), while both are outperformed by the proposed RU. 
	Despite its low performance with respect to class-agnostic measures, ED systematically achieves second best results with respect to mean Top-5 recall and mean $TtA(5)$. This highlights that there is no clear second-best performing method. Finally, mean $TtA(k)$ highlights that the proposed method can anticipate verbs, nouns and actions $1.62$, $1.11$ and $0.76$ seconds in advance respectively.

	\begin{table}[t]
		\begin{adjustbox}{width=\linewidth,center}
			\begin{tabular}{llccc|ccc}
				\hline
				& & \multicolumn{3}{c|}{Top-1 Accuracy\%} & \multicolumn{3}{c}{Top-5 Accuracy\%}  \\ \hline
				& & VERB & NOUN & ACT. & VERB & NOUN & ACT. \\ \hline
				\multirow{4}{*}{\rotatebox{90}{\textbf{S1}}} &
				2SCNN~\cite{Damen2018EPICKITCHENS} & \underline{29.76} & 15.15 & 04.32 & 76.03 & 38.56 & 15.21 \\
				&ATSN~\cite{Damen2018EPICKITCHENS} & 31.81 & \underline{16.22} & 06.00 & \underline{76.56} & \underline{42.15} & \underline{28.21} \\
				&MCE~\cite{furnari2018Leveraging} & 27.92 & 16.09 & \underline{10.76} & 73.59 & 39.32 & 25.28 \\
				&\textbf{RU} & \textbf{33.04} & \textbf{22.78} & \textbf{14.39} & \textbf{79.55} & \textbf{50.95} & \textbf{33.73} \\ \hline
				&Improvement & +1.23 & +6.56 & +3.63 & +2.99 & +8.80 & +5.52 \\

				\hline \hline
				\multirow{4}{*}{\rotatebox{90}{\textbf{S2}}} &
				2SCNN~\cite{Damen2018EPICKITCHENS} & 25.23 & 09.97 & 02.29 & \underline{68.66} & 27.38 & 09.35 \\
				&ATSN~\cite{Damen2018EPICKITCHENS} & \underline{25.30} & \underline{10.41} & 02.39 & 68.32 & \underline{29.50} & 06.63 \\
				&MCE~\cite{furnari2018Leveraging} & 21.27 & 09.90 & \underline{05.57} & 63.33 & 25.50 & \underline{15.71} \\
				&\textbf{RU} & \textbf{27.01} & \textbf{15.19} & \textbf{08.16} & \textbf{69.55} & \textbf{34.38} & \textbf{21.10} \\ \hline
				&Improvement & +1.71 & +4.78 & +2.59 & +0.89 & +4.88 & +5.39 \\

				\hline

			\end{tabular}
		\end{adjustbox}	\caption{Anticipation results on the EPIC-KITCHENS test sets.}
		\label{tab:anticipation_ek_test}
		\vspace{-\topsep}
	\end{table}

	\tablename~\ref{tab:anticipation_ek_test} assesses the performance of the proposed method on the official test sets of EPIC-Kitchens dataset.\footnote{See supp. for precision and recall results.} RU outperforms all competitors by consistent margins on both the ``seen'' test, which includes scenes appearing in the training set (\textbf{S1}) and the ``unseen'' test set, with scenes not appearing in the training set (\textbf{S2}). Also in this case, RU is strong on nouns, obtaining $+6.56\%$ and $+8.8\%$ in \textbf{S1}, as well as $+4.78\%$ and $+4.88$ in \textbf{S2}. Improvements in terms of actions are also significant: $+3.63\%$ and $+5.52\%$ in \textbf{S1}, as well as $+2.59\%$ and $+5.39\%$ on \textbf{S2}. 

	\vspace{\topsep}
	\noindent
	\textbf{Anticipation Results on EGTEA Gaze+\hspace{0.5em}}
	\begin{table}
		\begin{adjustbox}{width=\linewidth,center}
			\begin{tabular}{p{1.2cm}ccccccccc}
				\hline
				
				\multicolumn{1}{c}{} & \multicolumn{8}{c}{Top-5 ACTION Accuracy\% @ different $\tau_a$(s)}   & $TtA$ \\  \hline %
				\multicolumn{1}{c}{} & $2$ & $1.75$ & $1.5$ & $1.25$ & $1.0$& $0.75$&$0.5$&$0.25$& \\ \hline
				DMR~\cite{vondrick2016anticipating} & /              & /              & /              & /              & 55.70          & /              & /              & /              &  /\\
				ATSN~\cite{Damen2018EPICKITCHENS}   & /              & /              & /              & /              & 40.53          & /              & /              & /              &  / \\
				MCE~\cite{furnari2018Leveraging}    & /              & /              & /              & /              & 56.29          & /              & /              & /              & / \\
				ED~\cite{gao2017red}                & 45.03          & 46.22          & 46.86          & 48.36          & 50.22          & 51.86          & 49.99          & 49.17          & 1.24\\
				FN~\cite{de2018modeling}            & 54.06          & 54.94          & 56.75          & 58.34          & 60.12          & 62.03          & 63.96          & 66.45          & 1.26\\
				RL~\cite{ma2016learning}            & 55.18          & 56.31          & 58.22          & 60.35          & 62.56          & 64.65          & 67.35          & 70.42          & 1.29\\
				EL~\cite{jain2016recurrent}         & \underline{55.62} & \underline{57.56} & \underline{59.77} & \underline{61.58} & \underline{64.62} & \underline{66.89} & \underline{69.60} & \underline{72.38} & \underline{1.32}\\
				\textbf{RU}                                  & \textbf{56.82} & \textbf{59.13} & \textbf{61.42} & \textbf{63.53} & \textbf{66.40} & \textbf{68.41} & \textbf{71.84} & \textbf{74.28} & \textbf{1.41} \\ \hline
				Imp                                 & +1.20          & +1.57          & +1.65          & +1.95          & +1.78          & +1.52          & +2.24          & +1.89          & +0.09\\ \hline
			\end{tabular}
		\end{adjustbox}
		\caption{Anticipation results on EGTEA Gaze+.}
		\label{tab:anticipation_egtea}
		\vspace{-\topsep}
	\end{table} \tablename~\ref{tab:anticipation_egtea} reports Top-5 action accuracy scores achieved by the compared methods on EGTEA Gaze+ with respect to the $8$ considered anticipation times. The table also reports mean $TtA(5)$ action scores, denoted as $TtA$. The proposed method outperforms all competitors at all anticipation times. Note that the margins of improvement are smaller on EGTEA Gaze+ due to its smaller-scale ($106$ actions in vs $2,513$ actions in EPIC-KITCHENS). Differently from \tablename~{\ref{tab:anticipation_ek}}, EL systematically achieves second best performance on EGTEA Gaze+, which highlights again that there is no clear second best competitor to RU in our evaluations.
	
	\vspace{\topsep}
	\noindent
	\textbf{Ablation Study on EPIC-Kitchens\hspace{0.5em}}
	To assess the role of rolling-unrolling, we consider a strong baseline composed of a single LSTM (same configuration as R-LSTM) and three branches (RGB, Flow, OBJ) with late fusion (BL). To tear apart the influence of rolling-unrolling and MATT, \tablename~\ref{tab:ablation}(a) compares this baseline with the proposed RU architecture, where MATT has been replaced with late fusion. The proposed RU approach brings systematic improvements over the baseline for all anticipation times, with larger improvements in the case of the object branch.

\begin{figure}
		\includegraphics[width=\linewidth]{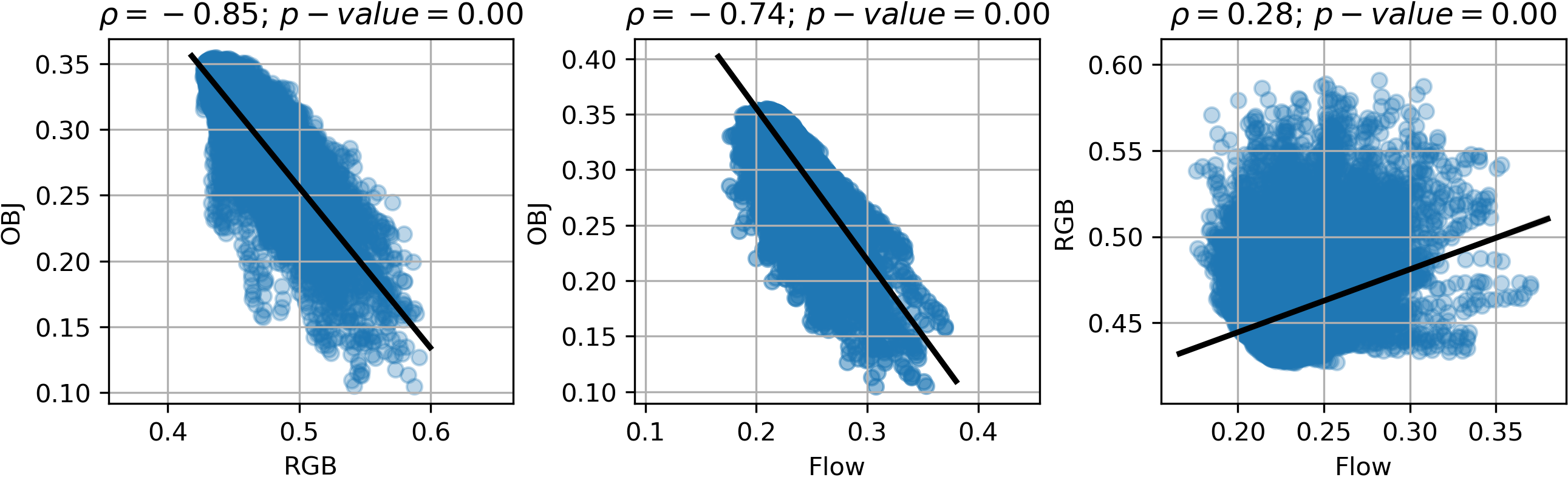}
		\caption{Correlations between modality attention weights}
		\label{fig:correlations}
		\vspace{-1.5\topsep}
	\end{figure}

	\tablename~\ref{tab:ablation}(b) reports the performances of the single branches of RU and compares MATT with respect to late fusion (i.e., averaging predictions) and early fusion (i.e., feeding the model with the concatenation of the modality-specific representations). MATT always outperforms late fusion, which consistently achieves second best results. Early fusion always leads to sub-optimal results. All fusion schemes always improve over the single branches. 
	\figurename~\ref{fig:correlations} shows regression plots of modality attention weights computed on all samples of the validation set. RGB and OBJ weights are characterized by a strong and steep correlation. A similar pattern applies to Flow and OBJ weights, whereas Flow and RGB weights are characterized by a small positive correlation. This suggests that MATT gives more credit to OBJ when RGB and Flow are less informative, whereas the it relies on RGB and Flow when objects are not necessary.

	\tablename~\ref{tab:ablation}(c) assesses the role of pre-training the proposed architecture with sequence completion. As can be noted, the proposed pre-training procedure brings small but consistent improvements for most anticipation times. \tablename~\ref{tab:ablation}(d) compares RU with the strong baseline of \tablename~\ref{tab:ablation}(a). The comparison highlights the cumulative effect of all the proposed procedures/component with respect to a strong baseline using three modalities. It is worth noting that the proposed architecture brings improvements for all anticipation times, ranging from $+1.53\%$ to $+4.08\%$.

	\begin{table}[t]
		\setlength{\tabcolsep}{5pt}
		\centering
		\begin{adjustbox}{width=\linewidth,center}
			\begin{tabular}{p{2cm}ccccccccc}
				\hline
				\multicolumn{1}{c}{} & \multicolumn{8}{c}{Top-5 ACTION Accuracy\% @ different $\tau_a$(s)} & $TtA$ \\  \hline %
				\multicolumn{1}{c}{} & $2$ & $1.75$ & $1.5$ & $1.25$ & $1.0$& $0.75$&$0.5$&$0.25$ & \\ \hline
				BL (Late)            & \underline{27.96} & \underline{28.76} & \underline{29.99} & \underline{31.09} & \underline{32.02} & \underline{33.09} & \underline{34.13} & \underline{34.92} & \underline{0.66}\\
				RU (Late)            & \textbf{29.10} & \textbf{29.77} & \textbf{31.72} & \textbf{33.09} & \textbf{34.23} & \textbf{35.28} & \textbf{36.10} & \textbf{37.61} & \textbf{0.73} \\ \hline
				Imp.                 & +1.14          & +1.01          & +1.73          & +2.00          & +2.21          & +2.19          & +1.97          & +2.69          & +0.07 \\ \hline
				& 
			\end{tabular}
		\end{adjustbox}
		
		\vspace{1mm}
		\centerline{\footnotesize(a) Rolling-Unrolling Mechanism.}
		\vspace{1mm}
		
		\begin{adjustbox}{width=\linewidth,center}
			\begin{tabular}{p{2cm}ccccccccc}
				\hline
				RU (RGB)             & 25.44 & {26.89} & {28.32} & {29.42} & {30.83} & {32.00} & {33.31} & {34.47} & 0.69\\ 
				RU (Flow)            & {17.38} & {18.04} & {18.91} & {19.97} & {21.42} & {22.37} & {23.49} & {24.18} & 0.51 \\ 
				RU (OBJ)             & {24.56} & {25.60} & {26.61} & {28.32} & {29.89} & {30.85} & {31.82} & {33.39} & 0.67 \\ \hline
				
				Early Fusion         & 25.58          & 27.25          & 28.58          & 29.59          & 31.88          & 32.78          & 33.99          & 35.62          & 0.72 \\
				Late Fusion          & \underline{29.10} & \underline{29.77} & \underline{31.72} & \underline{33.09} & \underline{34.23} & \underline{35.28} & \underline{36.10} & \underline{37.61} & \underline{0.73} \\
				MATT                 & \textbf{29.44} & \textbf{30.73} & \textbf{32.24} & \textbf{33.41} & \textbf{35.32} & \textbf{36.34} & \textbf{37.37} & \textbf{38.98} & \textbf{0.76} \\ \hline
				Imp.                 & +0.34          & +0.96          & +0.52          & +0.32          & +1.09          & +1.06          & +1.27          & +1.37          & +0.03\\
				\hline& 
			\end{tabular}
		\end{adjustbox}
		
		\vspace{1mm}
		\centerline{\footnotesize (b) Modality Attention Fusion Mechanism.}
		\vspace{1mm}
		
		\begin{adjustbox}{width=\linewidth,center}
			\begin{tabular}{p{2cm}ccccccccc}
				\hline
				w/o SCP              & \underline{29.22} & \underline{30.43} & \textbf{32.34} & \underline{33.37} & \underline{34.75} & \underline{35.84} & \underline{36.79} & \underline{37.93} & \underline{0.75} \\
				with SCP             & \textbf{29.44} & \textbf{30.73} & \underline{32.24} & \textbf{33.41} & \textbf{35.32} & \textbf{36.34} & \textbf{37.37} & \textbf{38.98} & \textbf{0.76} \\ \hline
				Imp. of SCP          & +0.22          & +0.30          & -0.10          & +0.04          & +0.57          & +0.50          & +0.58          & +1.05          & +0.01 \\ \hline
			\end{tabular}
		\end{adjustbox}
		
		\vspace{1mm}
		\centerline{\footnotesize (c) Sequence-Completion Pre-training.}

		\begin{adjustbox}{width=\linewidth,center}
			\begin{tabular}{p{2cm}ccccccccc}
				\hline
				BL (Fusion)          & \underline{27.96} & \underline{28.76} & \underline{29.99} & \underline{31.09} & \underline{32.02} & \underline{33.09} & \underline{34.13} & \underline{34.92} & \underline{0.66} \\
				RU (Fusion)          & \textbf{29.44} & \textbf{30.73} & \textbf{32.24} & \textbf{33.41} & \textbf{35.32} & \textbf{36.34} & \textbf{37.37} & \textbf{38.98} & \textbf{0.76} \\ \hline
				Imp. (Fusion)        & +1.48          & +1.97          & +2.25          & +2.32          & +3.30          & +3.25          & +3.24          & +4.06          & +0.1 \\ \hline
			\end{tabular}
		\end{adjustbox}
		
		\vspace{1mm}
		\centerline{\footnotesize (d) Overall comparison wrt strong baseline.}
		
		\caption{Ablation study on EPIC-KITCHENS.}
		\label{tab:ablation}
		\vspace{-\topsep}
	\end{table} 
	
\begin{figure*}
		\begin{tabular}{c}
			\includegraphics[width=\linewidth,clip=true,trim=0 0 0 90]{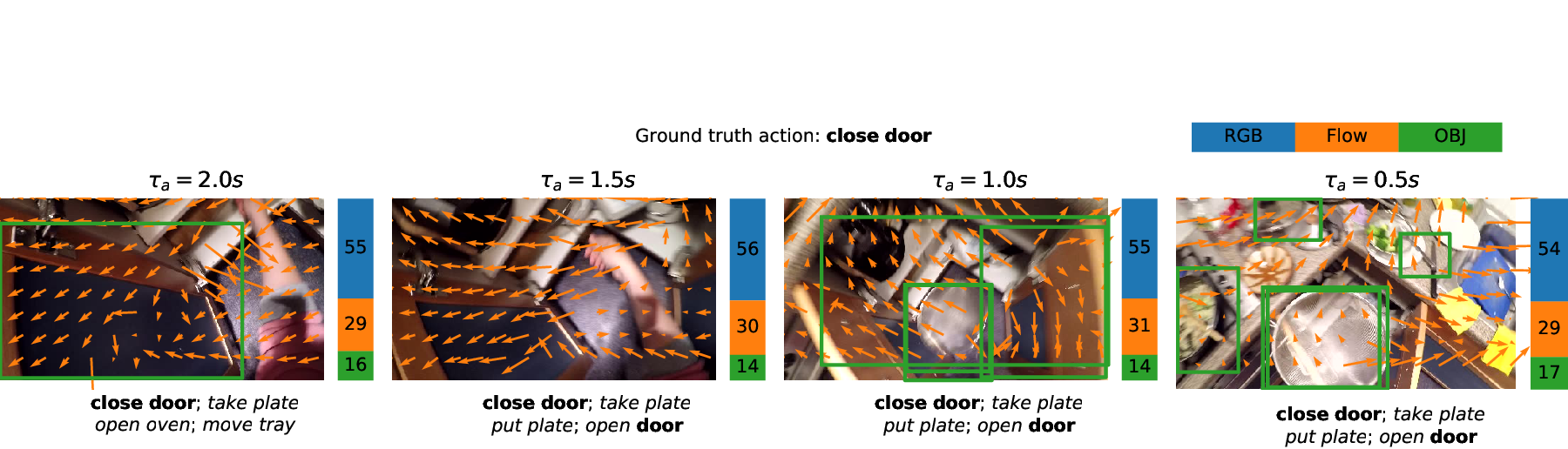}\\
			\includegraphics[width=\linewidth,clip=true,trim=0 0 0 110]{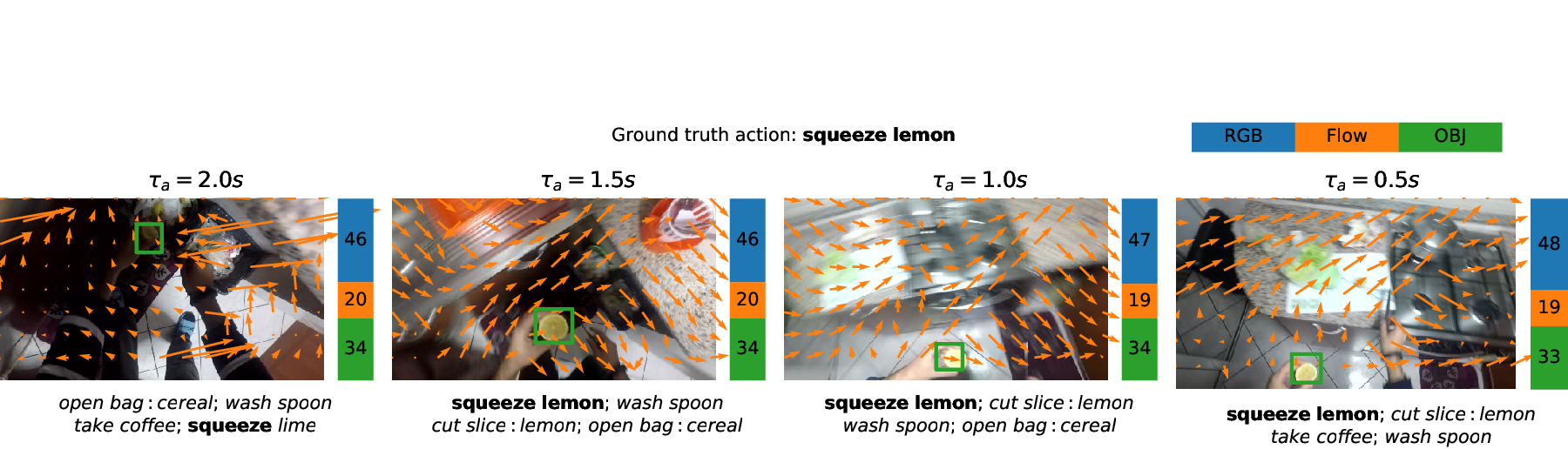}
		\end{tabular}
		\caption{Qualitative examples (best seen on screen). Legend for attention weights: blue - RGB, orange - Flow, green - objects. }
		\label{fig:qualitative}
		\vspace{-\topsep}
	\end{figure*}

	\vspace{\topsep}
	\noindent
	\textbf{Qualitative Results\hspace{0.5em}}
	\figurename~\ref{fig:qualitative} reports two examples of anticipations made by the proposed method at four anticipation times. Under each frame we report top-4 predictions, whereas modality weights computed by MATT are reported in percentage on the right. Green bounding boxes are shown around the detected objects and the optical flow is illustrated in orange. In the first example (top), the model can predict ``close door'' based on context and past actions (e.g., taking objects out of the cupboard), hence it assigns large weights to RGB and Flow and low weights to OBJ. In the second example (bottom), the model initially predicts ``squeeze lime'' at $\tau_a=2s$. Later, the prediction is corrected to ``squeeze lemon'' as soon as the lemon can be reliably detected. Note that in this case the network assigns larger weights to OBJ as compared to the previous example.\footnote{See supp. and \textit{https://iplab.dmi.unict.it/rulstm/} for additional examples and videos.}
	
	\vspace{\topsep}
	\noindent
	\textbf{Early Action Recognition\hspace{0.5em}}
	We also observe that the proposed method generalizes to the task of early action recognition. We adapt our processing scheme by sampling $8$ video snippets from each action segment uniformly in time and set $S_{enc}=0$, $S_{ant}=8$. The snippets are fed to the model, which produces predictions at each time-step, corresponding to the following observation rates:  $12.5\%$, $25\%$, $37.5\%$, $50\%$, $62.5\%$, $75\%$, $87.5\%$, $100\%$. Branches are fused by late fusion in this case. We compare the proposed method with respect to FN, EL and RL.
	\begin{table}[t]
		\begin{adjustbox}{width=\linewidth,center}
			\begin{tabular}{p{0.8cm}cccccccc}
				\multicolumn{1}{c}{} & \multicolumn{8}{c}{Top-1 ACTION Accuracy\% @ different observation rates} \\ \hline
				& $12.5\%$ & $25.0\%$ & $37.5\%$ & $50.0\%$ & $62.5\%$& $75.0\%$&$87.5\%$&$100\%$ \\
				\hline

				FN~\cite{de2018modeling}& 19.61 & 23.85 & 25.66 & 26.85 & 27.47 & 28.34 & 28.26 & 28.38\\
				EL~\cite{jain2016recurrent}& 19.69 & 23.27 & 26.03 & 27.49 & 29.06 & 29.97 & \underline{30.91} & \underline{31.46}\\
				RL~\cite{ma2016learning} & \underline{22.53} & \underline{25.08} & \underline{27.19} & \underline{28.64} & \underline{29.57} & \underline{30.13} & 30.45 & 30.47\\
				\textbf{RU} & \textbf{24.48} & \textbf{27.63} & \textbf{29.44} & \textbf{30.93} & \textbf{32.16} & \textbf{33.09} & \textbf{33.63} & \textbf{34.07}\\
				
				\hline
				Imp.&+1.95 & +2.55 & +2.25 & +2.29 & +2.59 & +2.96 & +2.72 & +2.61\\
				
				\hline
			\end{tabular}
		\end{adjustbox}
		\caption{Early recognition results on EPIC-KITCHENS.}
		\label{tab:recognition_ek}
		\vspace{-1.5\topsep}
	\end{table} %
	\begin{table}[t]
		\begin{adjustbox}{width=\linewidth,center}
			\begin{tabular}{p{0.8cm}cccccccc}
				\multicolumn{1}{c}{} & \multicolumn{8}{c}{Top-1 ACTION Accuracy\% @ different observation rates} \\ \hline
				& $12.5\%$ & $25.0\%$ & $37.5\%$ & $50.0\%$ & $62.5\%$& $75.0\%$&$87.5\%$&$100\%$ \\ \hline
				FN~\cite{de2018modeling}    & 44.02          & 50.32          & 53.34          & 55.10          & 56.58          & 57.31          & 57.95          & 57.72          \\
				EL~\cite{jain2016recurrent} & 40.31          & 48.08          & 51.84          & 54.71          & 56.93          & 58.45          & \underline{59.55} & \underline{60.18} \\
				RL~\cite{ma2016learning}    & \underline{45.42} & \underline{51.00} & \underline{54.20} & \underline{56.54} & \underline{58.09} & \underline{58.93} & 59.29          & 59.50          \\
				\textbf{RU}                          & \textbf{45.94} & \textbf{51.84} & \textbf{54.39} & \textbf{57.05} & \textbf{58.15} & \textbf{59.31} & \textbf{60.10} & \textbf{60.20} \\
				\hline
				Imp.                        & +0.51          & +0.84          & +0.20          & +0.51          & +0.06          & +0.38          & +0.55          & +0.02         \\
				\hline
			\end{tabular}
		\end{adjustbox}
		\caption{Early recognition results on EGTEA Gaze+.}
		\label{tab:recognition_egtea}
		\vspace{-1.5\topsep}
	\end{table} \tablename~\ref{tab:recognition_ek} reports Top-1 early action recognition accuracy results obtained by the compared methods on our validation set of EPIC-Kitchens. The proposed method consistently outperforms the competitors at all observation rates. Interestingly, RU achieves an early action recognition accuracy of $33.09\%$ when observing only $75\%$ of the action, which is already comparable to the accuracy of $34.07\%$ achieved when the full action is observed. This indicates that RU can timely recognize actions before they are completed. RL achieves second best results up to observation rate $75\%$, whereas EL achieves second best results when more than $75\%$ of the action is observed, which confirms the lack of a clear second-best performer.
	\tablename~\ref{tab:recognition_egtea} reports Top-1 accuracy results obtained on EGTEA Gaze+. The proposed RU outperforms the competitors for all observation rates by small but consistent margins. Coherently with \tablename~\ref{tab:recognition_ek}, second best results are obtained by RL and EL.

	\begin{table}[t]
		\begin{adjustbox}{width=\linewidth,center}
			\begin{tabular}{llccc|ccc}
				\hline
				& & \multicolumn{3}{c|}{Top-1 Accuracy\%} & \multicolumn{3}{c}{Top-5 Accuracy\%}  \\ \hline
				& & VERB & NOUN & ACTION & VERB & NOUN & ACTION \\ \hline
				\multirow{5}{*}{\rotatebox{90}{\textbf{S1}}} &
				2SCNN~\cite{Damen2018EPICKITCHENS} & 42.16 & 29.14 & 13.23 & 80.58 & 53.70 & 30.36\\
				&TSN~\cite{Damen2018EPICKITCHENS} & 48.23 & 36.71 & 20.54 & 84.09 & \underline{62.32} & 39.79\\
				&LSTA~\cite{sudhakaran2018lsta} & \textbf{59.55} & 38.35 & \underline{30.33} & \textbf{85.77} & 61.49 & \underline{49.97}\\
				&MCE~\cite{furnari2018Leveraging} & 54.22 & \underline{38.85} & 29.00 & 85.22 & 61.80 & 49.62\\
				&\textbf{RU} & \underline{56.93} & \textbf{43.05} & \textbf{33.06} & \underline{85.68} & \textbf{67.12} & \textbf{55.32} \\ \hline
				
				&Imp. & -2.62 & +4.20 & +2.73 & -0.09 & +4.80 & +5.35\\
				
				\hline
				\multirow{5}{*}{\rotatebox{90}{\textbf{S2}}} &
				2SCNN~\cite{Damen2018EPICKITCHENS} & 36.16 & 18.03 & 07.31 & 71.97 & 38.41 & 19.49\\
				&TSN~\cite{Damen2018EPICKITCHENS} & 39.40 & 22.70 & 10.89 & \underline{74.29} & \underline{45.72} & 25.26\\
				&LSTA~\cite{sudhakaran2018lsta} & \textbf{47.32} & 22.16 & \underline{16.63} & \textbf{77.02} & 43.15 & 30.93 \\
				&MCE~\cite{furnari2018Leveraging} & 40.90 & \underline{23.46} & 16.39 & 72.11 & 43.05 & \underline{31.34} \\
				&\textbf{RU} & \underline{43.67} & \textbf{26.77} & \textbf{19.49} & 73.30 & \textbf{48.28} & \textbf{37.15} \\ \hline
				& Imp. & -3.65 & +3.31 & +2.86 & -3.72 & +2.56 & +5.81 \\
				
				\hline

			\end{tabular}
		\end{adjustbox}	\caption{Recognition results on the EPIC-KITCHENS test sets.}
		\label{tab:recognition_ek_test}
		\vspace{-\topsep}
	\end{table}

	\vspace{\topsep}
	\noindent
	\textbf{Egocentric Action Recognition Results\hspace{0.5em}}
	The proposed method can be used to perform egocentric action recognition by considering the predictions obtained for the observation rate of $100\%$. \tablename~\ref{tab:recognition_ek_test} compares the performance of the proposed method with other egocentric action recognition methods on the two test sets of EPIC-Kitchens.\footnote{See supp. for precision and recall results.} Our RU outperforms all competitors in recognizing actions and nouns on both sets by significant margins, whereas it achieves second-best results in most cases for verb recognition. RU obtains an action recognition accuracy of $60.2\%$ on EGTEA Gaze+. Despite being designed for action anticipation, RU outperforms recent approaches, such as Li et al.~\cite{Li_2018_ECCV} ($+6.9\%$ wrt $53.3\%$) and Zhang et al.~\cite{zhang2018adding} ($+3.19\%$ wrt $57.01\%$ - reported from~\cite{sudhakaran2018lsta}), and obtaining performances comparable to state-of-the-art approaches such as Sudhakaran and Lanz~\cite{sudhakaran2018attention} ($-0.56\%$ wrt $60.76$) and Sudhakaran et al.~\cite{sudhakaran2018lsta} ($-1.66\%$ wrt $61.86\%$).\footnote{See supp. for the full table.}

	\vspace{-\topsep}
	\section{Conclusion}
	\vspace{-0.5\topsep}
	We presented RU-LSTM, a learning architecture which processes RGB, optical flow and object-based features using two LSTMs and a modality attention mechanism to anticipate actions from egocentric video. Experiments on two datasets show the superiority of the approach with respect to prior art and highlight generalization over datasets and tasks: anticipation, early recognition, and recognition. %
	
	\vspace{-\topsep}
	\section*{Acknowledgment}
	\vspace{-0.5\topsep}
	This research is supported by Piano della Ricerca 2016-2018, linea di
	Intervento 2 of DMI, University of Catania.

\appendix

\section{Implementation Details and Training Procedure of the Proposed Method}
\label{sec:implementation_details}
\begin{figure*}
	\includegraphics[width=\linewidth]{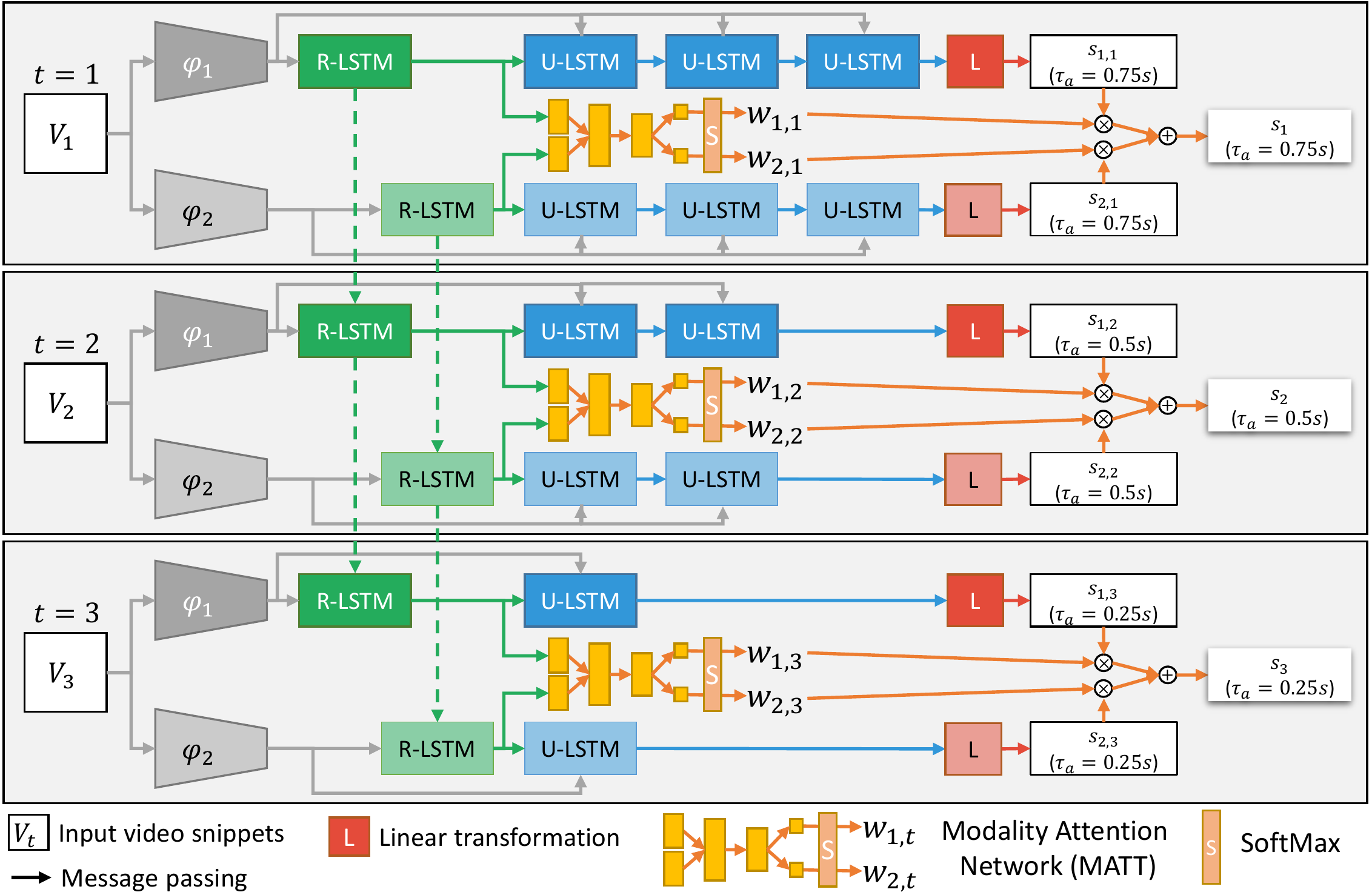}
	\label{fig:architecture}
	\caption{Example of the proposed architecture with $M=2$ modalities. In our experiments, we use three modalities: RGB, Flow and OBJ. Modules belonging to different branches are illustrated using different color shades.}
\end{figure*}

This section reports the implementation and training details of both the proposed and compared methods. A diagram of our architecture is reported in \figurename~\ref{fig:architecture} for the convenience of the reader. The reader is referred to the paper for a description of the architecture. 

\subsection{Architectural Details of RU-LSTM and MATT}
We use a Batch Normalized Inception CNN~\cite{ioffe2015batch} (BNInception) in the spatial and flow branches and consider the $1024$-dimensional vectors produced by the last global average pooling layer of the network as output representations. Optical flows are extracted using the TVL1 algorithm~\cite{zach2007duality}. Specifically, we use the pre-computed optical flows provided by the authors in the case of EPIC-Kitchens (see \url{http://EPIC-Kitchens.github.io/}) and compute optical flows on EGTEA Gaze+ using the code provided in \url{https://github.com/feichtenhofer/gpu_flow} with default parameters. At test time, the CNNs are fed with input images and optical flows resized to $456 \times 256$ pixels. Note that, due to global average pooling, the output of the BNInception CNN will be a $1024$ feature vector regardless size of the input image. We found this setting leading to better performance as compared to extracting a $224 \times 224$ crop from the center of the image. For the object branch, we use a Faster R-CNN object detector~\cite{girshick2015fast} with a ResNet-101 backbone~\cite{he2016deep}, as implemented in~\cite{Detectron2018}. Both the Rolling LSTM (R-LSTM) and the Unrolling LSTM (U-LSTM) contain a single layer with $1024$ hidden units. Dropout with $p=0.8$ is applied to the input of each LSTM and to the input of the final fully connected layer used to obtain class scores. The Modality ATTention network (MATT) is a feed-forward network with three fully connected layers containing respectively $h/4$, $h/8$ and $3$ hidden units, where $h=6144$ is the dimension of the input to the attention network (i.e., the concatenation of the hidden and cell states of $1024$ units each related to the three R-LSTMs). Dropout with $p=0.8$ is applied to the input of the second and third layers of the attention network to avoid over-fitting. ReLU activation function are used within the attention network.

\subsection{Training Procedure of RU-LSTM and MATT}
While the proposed architecture could be in principle trained in an end-to-end fashion, we found it extremely challenging to avoid over-fitting during end-to-end training. This is mainly due to the indirect relationship between input video and future actions. Indeed, differently from \textit{action recognition}, where the objects and actions to be recognized are present or take place in the input video, in the case of \textit{action anticipation}, the system should be able to anticipate objects and actions which do not always appear in the input video, which makes it hard to learn good representations end-to-end. To avoid over-fitting, the proposed architecture is trained as follows. First, we independently train the spatial and motion CNNs for the task of egocentric action recognition within the framework of TSN~\cite{wang2016temporal}. Specifically, we set the number of segments to $3$ and train the TSN models with Stochastic Gradient Descent (SGD) using standard cross entropy for $160$ epochs with an initial learning rate equal to $0.001$, which is decreased by a factor of $10$ after $80$ epochs. We use a mini-batch size of $64$ samples and train the models on a single Titan X. For all other parameters, we use the values recommended in~\cite{wang2016temporal}. We train the object detector to recognize the $352$ object classes of the EPIC-Kitchens dataset. We use the same object detector trained on EPIC-Kitchens when performing experiments on EGTEA Gaze+, as the latter dataset does not contain object bounding box annotations. This training procedure allows to learn the parameters $\theta^1$, $\theta^2$ and $\theta^3$ of the representation functions related to the three modalities (i.e., RGB, Flow, OBJ). After this procedure, these parameters are fixed and they are no more optimized. For efficiency, we pre-compute representations over the whole dataset.

Each branch of the RU-LSTM is training with SGD using the cross entropy loss with a fixed learning rate equal to $0.01$ and momentum equal to $0.9$. Each branch is first pre-trained with Sequence Completion Pre-training (SCP). Specifically, appearance and motion branches are trained for $100$ epochs, whereas the object branch is trained for $200$ epochs. Branches are then fine-tuned for the action anticipation task. Once each branch has been trained, the complete architecture with three branches is assembled to form a three-branch network and the model is further fine-tuned for $100$ epochs using cross entropy and the same learning parameters. In \figurename~\ref{fig:architecture} an example of a two-branches architecture is shown.

In the case of early action recognition, each branch is trained for $200$ epochs (both SCP and main task) with a fixed learning rate equal to $0.01$ and momentum equal to $0.9$.

Note that, in order to improve performances, we apply early stopping at each training stage. This is done by choosing the iterations of the intermediate and final models which obtain the best Top-5 action anticipation accuracy for the anticipation time $\tau_a=1s$ on the validation set. In the case of \textit{early action recognition}, we choose the epoch obtaining the best average Top-1 action accuracy across observation rates. The same early stopping strategy is applied to all the methods for fair comparison. The proposed RU-LSTM architecture has been implemented using the PyTorch library~\cite{paszke2017automatic}. The code will be provided upon publication, together with all details and data useful to replicate the results.

\subsubsection{Note on End-To-End Training}
We chose to fix the feature extractors in our work as we experienced over-fitting when training the model end-to-end. Specifically, we tried the following: (1) Training the RGB branch end-to-end from scratch (except the CNN, which is pre-trained on Imagenet), (2) Pre-training the CNN on the action recognition task with TSN, then training the RGB branch end-to-end, (3) Training the RGB branch using fixed representations as described in the paper, then fine-tuning the CNN + RU-LSTM model end-to-end. In our experiments, (1) and (2) led to poor performance already at the sequence-completion stage, while (3) did not improve performance. Our insight is that the indirect relationship between the observed scene and the action yet to take place can make learning representations end-to-end much more difficult than in the case of recognition. For instance, when anticipating the action ``take cup'', the object ``cup'' may or may not be present in the observed video segment, which makes unclear what visual features the CNN should extract. 

\subsection{Inference At a Fixed Anticipation Time}
Our model makes multiple anticipations at time-steps $7-14$. Also, since the predictions are updated as the video is processed and more evidence is acquired, such predictions may indeed be inconsistent (with anticipations performed closer to the beginning of the action being more likely to be correct). However, it should be noted that each prediction is deemed to be specific to a given anticipation time. For instance, at time-step $11$, the model tries to anticipate actions happening in $1s$. Therefore, the proposed approach can be used to anticipate actions at a fixed anticipation time by processing the buffered video up to the related time-step, discarding all other predictions. E.g., if the anticipation time is set to $\tau_a=1s$, the model should process the last $11$ time-steps.

\subsection{Choice of Parameters $\alpha$, $S_{enc}$ and $S_{ant}$.\hspace{0.5em}}
In this work, we set $\alpha=0.25s$ and $S_{ant}=8$ to generalize the settings of the EPIC-Kitchens anticipation challenge. Indeed, in these settings, we can anticipate actions up to $2s$ in advance ($8 \times 0.25s$), while still being able to produce anticipations at anticipation time $\tau_a=1s$ ($4 \times 0.25s$) as required for the challenge. We investigated the effect of $S_{enc}$ when we fix $\alpha=0.25s$ and $S_{dec}=8$. 
We noted that the choice of $S_{enc}$ affects performance lightly and hence chose $S_{enc}=6$ to maximize action anticipation performance for anticipation time $\tau_a=1s$.

\section{Implementation Details of the Compared Methods}
\label{sec:implementation_details_compared}
Since no official public implementation is available for the compared methods, we performed experiments using our own implementations. In this section, we report the implementation details of each of the compared method.

\subsection{Deep Multimodal Regressor (DMR)}
We implement the Deep Multimodal Regressor proposed in~\cite{vondrick2016anticipating} setting the number of multi-modal branches with interleaved units to $k=3$. For fair comparisons, we substituted the AlexNet backbone originally considered in~\cite{vondrick2016anticipating} with a BNInception CNN pre-trained on ImageNet. The CNN is trained to anticipate future representations extracted using BNInception pre-trained on ImageNet using the procedure proposed by the authors. Specifically, we perform mode update every epoch. Since training an SVM with large number of classes is challenging (in our settings, we have $2,513$ different action classes), we substituted the SVM with a Multi Layer Perceptron (MLP) with $1024$ hidden units and dropout with $p=0.8$ applied to the input of the first and second layer. To comply with the pipeline proposed in~\cite{vondrick2016anticipating}, we pre-train the model on our training split of EPIC-Kitchens in an unsupervised fashion and train the MLP separately on representations pre-extracted from the training set using the optimal modes found at training time. As a result, during the training of the MLP, the weights of the CNN are not optimized. The DMR architecture is trained with Stochastic Gradient Descent using a fixed learning rate equal to $0.1$ and a momentum equal to $0.9$. The network is trained for several epochs until the validation loss saturates. Note that training the CNN on the EPIC-Kitchens dataset takes several days on a single Titan X GPU using our implementation. After training, we apply early stopping by selecting the iteration with the lowest validation loss. The MLP is then trained with Stochastic Gradient Descent with fixed learning rate equal to $0.01$ and momentum equal to $0.9$. Early stopping is applied also in this case considering the iteration of the MLP achieving the highest Top-5 action accuracy on the validation set.

\subsection{Anticipation TSN (ATSN)}
We implement this model considering the TSN architecture used to pre-train the CNNs employed in the RGB and Flow branches of our architecture. We modify the network to output verb and noun scores and train it summing the cross entropy losses applied independently to verbs and nouns as specified in~\cite{Damen2018EPICKITCHENS}. At test time, we obtain action probabilities by assuming independence of verbs and nouns as follows: $p(a=(v,n)|x)=p(v|x)\cdot p(n|x)$, where $a=(v,n)$ is an action involving verb $v$ and noun $n$, $x$ is the input sample, whereas $p(v|x)$ and $p(n|x)$ are the probabilities computed directly by the network.

\subsection{ATSN + VNMCE Loss (MCE)}
This method is implemented training the TSN architecture used for ATSN with the Verb-Noun Marginal Cross Entropy Loss proposed in~\cite{furnari2018Leveraging}. We used the official code provided by the authors (\url{https://github.com/fpv-iplab/action-anticipation-losses/}).

\subsection{Encoder-Decoder LSTM (ED)}
We implement this model following the details specified in~\cite{gao2017red}. For fair comparison with respect to the proposed method, the model takes RGB and Flow features obtained using the representation functions as input for the RGB and Flow modalities used in our RU architecture. Differently from~\cite{gao2017red}, we do not include a reinforcement learning term in the loss as our aim is not to distinguish the action from the background as early as possible as proposed in~\cite{gao2017red}. The hidden state of the LSTMs is set to $2048$ units. The model encodes representations for $20$ steps, while decoding is carried out for $10$ steps at a step-size of $0.25s$. The architecture is trained on top of pre-extracted representations for $100$ epochs with the Adam optimizer and a learning rate of $0.001$. 

\subsection{Feedback-Network LSTM (FN)}
The method proposed in~\cite{de2018modeling} has been implemented considering the most performing architecture investigated by the authors, which comprises the ``optional'' LSTM layer and performs fusion by concatenation. The network uses our proposed video processing strategy. For fair comparison, we implement the network as a two-stream architecture with two branches processing independently RGB and Flow features. Final predictions are obtained with late fusion (equal weights for the two modalities). We use the representation functions of our architecture to obtain RGB and Flow features. The model has hidden layers of $1024$ units, which in our experiments leaded to improved results with respect to the $128$ features proposed by the authors~\cite{de2018modeling}. The model is trained using the same parameters used in the proposed architecture.

\subsection{RL \& EL}
These two methods are implemented considering a single LSTM with the same parameters of our Rolling LSTM. Similarly to FN, the models are trained as two-stream models with late fusion used to obtain final predictions. The input RGB and Flow features are computed using the representation functions considered in our architecture. The models are trained with the same parameters used in the proposed architecture. RL is trained using the ranking loss on the detection score proposed in~\cite{ma2016learning}, whereas EL is trained using the exponential anticipation loss proposed in~\cite{jain2016recurrent}.

\section{Additional Results}
\label{sec:results}
This section reports the full set of anticipation and recognition results on EPIC-Kitchens, including precision and recall, as well as the full table of comparisons of the proposed method on EGTEA Gaze+ for action recognition.

\begin{table*}[t]
	\begin{adjustbox}{width=\linewidth,center}
		\setlength{\tabcolsep}{3pt}
		\begin{tabular}{llccc|ccc|ccc|ccc}
			& & \multicolumn{3}{c|}{Top-1 Accuracy\%} & \multicolumn{3}{c|}{Top-5 Accuracy\%} & \multicolumn{3}{c|}{Avg Class Precision\%} & \multicolumn{3}{c}{Avg Class Recall\%} \\ \hline
			& & VERB & NOUN & ACTION & VERB & NOUN & ACTION & VERB & NOUN & ACTION & VERB & NOUN & ACTION \\ \hline
			\multirow{4}{*}{\rotatebox{90}{\textbf{S1}}} &
			2SCNN (Fusion)~\cite{Damen2018EPICKITCHENS} & \underline{29.76} & 15.15 & 04.32 & 76.03 & 38.56 & 15.21 & 13.76 & 17.19 & 02.48 & 07.32 & 10.72 & 01.81\\
			&TSN (Fusion)~\cite{Damen2018EPICKITCHENS} & 31.81 & \underline{16.22} & 06.00 & \underline{76.56} & \underline{42.15} & \underline{28.21} & \underline{23.91} & \underline{19.13} & 03.13 & 09.33 & \underline{11.93} & 02.39\\
			&VNMCE~\cite{furnari2018Leveraging} & 27.92 & 16.09 & \underline{10.76} & 73.59 & 39.32 & 25.28 & 23.43 & 17.53 & \underline{06.05} & \underline{14.79} & 11.65 & \underline{05.11}\\
			
			&RU-LSTM & \textbf{33.04} & \textbf{22.78} & \textbf{14.39} & \textbf{79.55} & \textbf{50.95} & \textbf{33.73} & \textbf{25.50} & \textbf{24.12} & \textbf{07.37} & \textbf{15.73} & \textbf{19.81} & \textbf{07.66}\\ \hline
			&Imp. wrt best & +1.23 & +6.56 & +3.63 & +2.99 & +8.80 & +5.52 & +1.59 & +4.99 & +1.32 & +0.94 & +7.88 & +2.55\\

			\hline
			\multirow{4}{*}{\rotatebox{90}{\textbf{S2}}} &
			2SCNN (Fusion)~\cite{Damen2018EPICKITCHENS} & 25.23 & 09.97 & 02.29 & \underline{68.66} & 27.38 & 09.35 & \textbf{16.37} & 06.98 & 00.85 & 05.80 & 06.37 & 01.14\\
			&TSN (Fusion)~\cite{Damen2018EPICKITCHENS} & \underline{25.30} & \underline{10.41} & 02.39 & 68.32 & \underline{29.50} & 06.63 & 07.63 & \underline{08.79} & 00.80 & 06.06 & \underline{06.74} & 01.07\\
			&VNMCE~\cite{furnari2018Leveraging} & 21.27 & 09.90 & \underline{05.57} & 63.33 & 25.50 & \underline{15.71} & 10.02 & 06.88 & \underline{01.99} & \underline{07.68} & 06.61 & \underline{02.39}\\
			
			&RU-LSTM & \textbf{27.01} & \textbf{15.19} & \textbf{08.16} & \textbf{69.55} & \textbf{34.38} & \textbf{21.10} & {\underline{13.69}} & \textbf{09.87} & \textbf{03.64} & \textbf{09.21} & \textbf{11.97} & \textbf{04.83}\\ \hline
			&Imp. wrt best & +1.71 & +4.78 & +2.59 & +0.89 & +4.88 & +5.39 & -2.68 & +1.08 & +1.65 & +1.53 & +5.23 & +2.44\\

			\hline

		\end{tabular}
	\end{adjustbox}	\caption{Egocentric action anticipation results on the EPIC-Kitchens test set.}
	\label{tab:anticipation_ek_test}
\end{table*}

\tablename~\ref{tab:anticipation_ek_test} compares the proposed method with respect to the competitors according to the full set of measures proposed along with the egocentric \textit{action anticipation} challenge~\cite{Damen2018EPICKITCHENS}, including precision and recall (which could not be included in the paper due to space limits). The proposed approach outperforms all competitors also according to precision and recall on \textbf{S1} and \textbf{S2}, except for average verb precision, where it is outperform by the two-stream CNN. Note that, coherently with Top-1 and Top-5 accuracy, the proposed method achieves large gains for noun precision and recall. Also note the small drop in performance between Top-1 noun accuracy and average noun recall on \textbf{S1} (from $22.78\%$ to $19.81\%$), which highlights balanced noun predictions.

\begin{table*}[t]
	\begin{adjustbox}{width=\linewidth,center}
		\setlength{\tabcolsep}{3pt}
		\begin{tabular}{llccc|ccc|ccc|ccc}
			& & \multicolumn{3}{c|}{Top-1 Accuracy\%} & \multicolumn{3}{c|}{Top-5 Accuracy\%} & \multicolumn{3}{c|}{Avg Class Precision\%} & \multicolumn{3}{c}{Avg Class Recall\%} \\ \hline
			& & VERB & NOUN & ACTION & VERB & NOUN & ACTION & VERB & NOUN & ACTION & VERB & NOUN & ACTION \\ \hline
			\multirow{5}{*}{\rotatebox{90}{\textbf{S1}}} &
			2SCNN (Fusion)~\cite{Damen2018EPICKITCHENS} & 42.16 & 29.14 & 13.23 & 80.58 & 53.70 & 30.36 & 29.39 & 30.73 & 05.53 & 14.83 & 21.10 & 04.46\\
			&TSN (Fusion)~\cite{Damen2018EPICKITCHENS} & 48.23 & 36.71 & 20.54 & 84.09 & \underline{62.32} & 39.79 & 47.26 & 35.42 & 10.46 & 22.33 & 30.53 & 08.83\\
			&LSTA~\cite{sudhakaran2018lsta} & \textbf{59.55} & 38.35 & \underline{30.33} & \textbf{85.77} & 61.49 & \underline{49.97} & 42.72 & 36.19 & 14.46 & \textbf{38.12} & \underline{36.19} & \underline{17.76}\\
			&VNMCE~\cite{furnari2018Leveraging} & 54.22 & \underline{38.85} & 29.00 & 85.22 & 61.80 & 49.62 & \textbf{53.87} & \underline{38.18} & \underline{18.22} & 35.88 & 32.27 & 16.56\\
			&RU-LSTM & \underline{56.93} & \textbf{43.05} & \textbf{33.06} & \underline{85.68} & \textbf{67.12} & \textbf{55.32} & \underline{50.42} & \textbf{39.84} & \textbf{18.91} & \underline{37.82} & \textbf{38.11} & \textbf{19.12}\\ \hline
			&Imp. & -2.62 & +4.20 & +2.73 & -0.09 & +4.80 & +5.35 & -3.45 & +1.66 & +0.69 & -0.30 & +1.92 & +1.36\\
			
			\hline
			\multirow{5}{*}{\rotatebox{90}{\textbf{S2}}} &
			2SCNN (Fusion)~\cite{Damen2018EPICKITCHENS} & 36.16 & 18.03 & 07.31 & 71.97 & 38.41 & 19.49 & 18.11 & 15.31 & 02.86 & 10.52 & 12.55 & 02.69\\
			&TSN (Fusion)~\cite{Damen2018EPICKITCHENS} & 39.40 & 22.70 & 10.89 & \underline{74.29} & \underline{45.72} & 25.26 & 22.54 & 15.33 & 05.60 & 13.06 & 17.52 & 05.81\\
			&LSTA~\cite{sudhakaran2018lsta} & \textbf{47.32} & 22.16 & \underline{16.63} & \textbf{77.02} & 43.15 & 30.93 & \textbf{31.57} & \underline{17.91} & \underline{08.97} & \textbf{26.17} & 17.80 & \underline{11.92}\\
			&VNMCE~\cite{furnari2018Leveraging} & 40.90 & \underline{23.46} & 16.39 & 72.11 & 43.05 & \underline{31.34} & \underline{26.62} & 16.83 & 07.10 & 15.56 & 17.70 & 10.17\\
			&RU-LSTM & \underline{43.67} & \textbf{26.77} & \textbf{19.49} & 73.30 & \textbf{48.28} & \textbf{37.15} & 23.40 & \textbf{20.82} & \textbf{09.72} & \underline{18.41} & \textbf{21.59} & \textbf{13.33}\\ \hline
			& Imp. & -3.65 & +3.31 & +2.86 & -3.72 & +2.56 & +5.81 & -8.17 & +2.91 & +0.75 & -7.76 & +3.79 & +1.41\\
			\hline

		\end{tabular}
	\end{adjustbox}
	\caption{Egocentric action recognition results on the EPIC-Kitchens test set.}
	\label{tab:recognition_ek_test}
\end{table*}

\tablename~\ref{tab:recognition_ek_test} compares the proposed method with respect to the competitors according to the full set of measures proposed with the \textit{egocentric action recognition} challenge~\cite{Damen2018EPICKITCHENS}. Similarly to \tablename~\ref{tab:anticipation_ek_test}, this includes precision and recall, which could not be included in the paper due to space limits. Similarly to what observed in the case of top-1 and top-5 accuracy, the proposed method outperforms the competitors according to most of the considered measures, despite not being explicitly designed to tackle the recognition task (i.e., our architecture was designed for the \textit{egocentric action anticipation} task.)

\tablename~\ref{tab:recognition_egtea_test} compares the proposed RU method against the state-of-the-art when tackling the task of egocentric action recognition on EGTEA Gaze+. It is worth noting that the proposed method outperforms many recent approaches by significant margins. It is also comparable with other state-of-the-art approaches such as the ones proposed in~\cite{sudhakaran2018lsta,sudhakaran2018attention}. Again, note that our architecture generalizes despite not being explicitly designed for the recognition task.

\begin{table}[t]
	\begin{adjustbox}{width=0.65\linewidth,center}
		\begin{tabular}{l|c|c}
			\hline
			Method & Acc.\% & Imp. \\ \hline
			Lit et al.~\cite{li2015delving} &	46.50 &	+13.7 \\
			Li et al.~\cite{Li_2018_ECCV} & 53.30	& +6.90\\
			Two stream~\cite{simonyan2014two}	& 41.84 &	+18.7\\
			I3D~\cite{carreira2017quo}	& 51.68 &	+8.52\\
			TSN~\cite{wang2016temporal}	& 55.93 &	+4.27\\
			eleGAtt~\cite{zhang2018adding}	& 57.01&	+3.19\\
			ego-rnn~\cite{sudhakaran2018attention}	& \underline{60.76}&	-0.56\\
			LSTA~\cite{sudhakaran2018lsta}	& \textbf{61.86}&	-1.66\\
			\textbf{RU}	& 60.20 & /	\\
			\hline
		\end{tabular}
	\end{adjustbox}
	\caption{Recognition results on EGTEA Gaze+.}
	\label{tab:recognition_egtea_test}
\end{table}

\section{EPIC-Kitchens Egocentric Action Anticipation Challenge Leaderboards}
\label{sec:challenge}
\begin{figure*}
	\includegraphics[width=\linewidth]{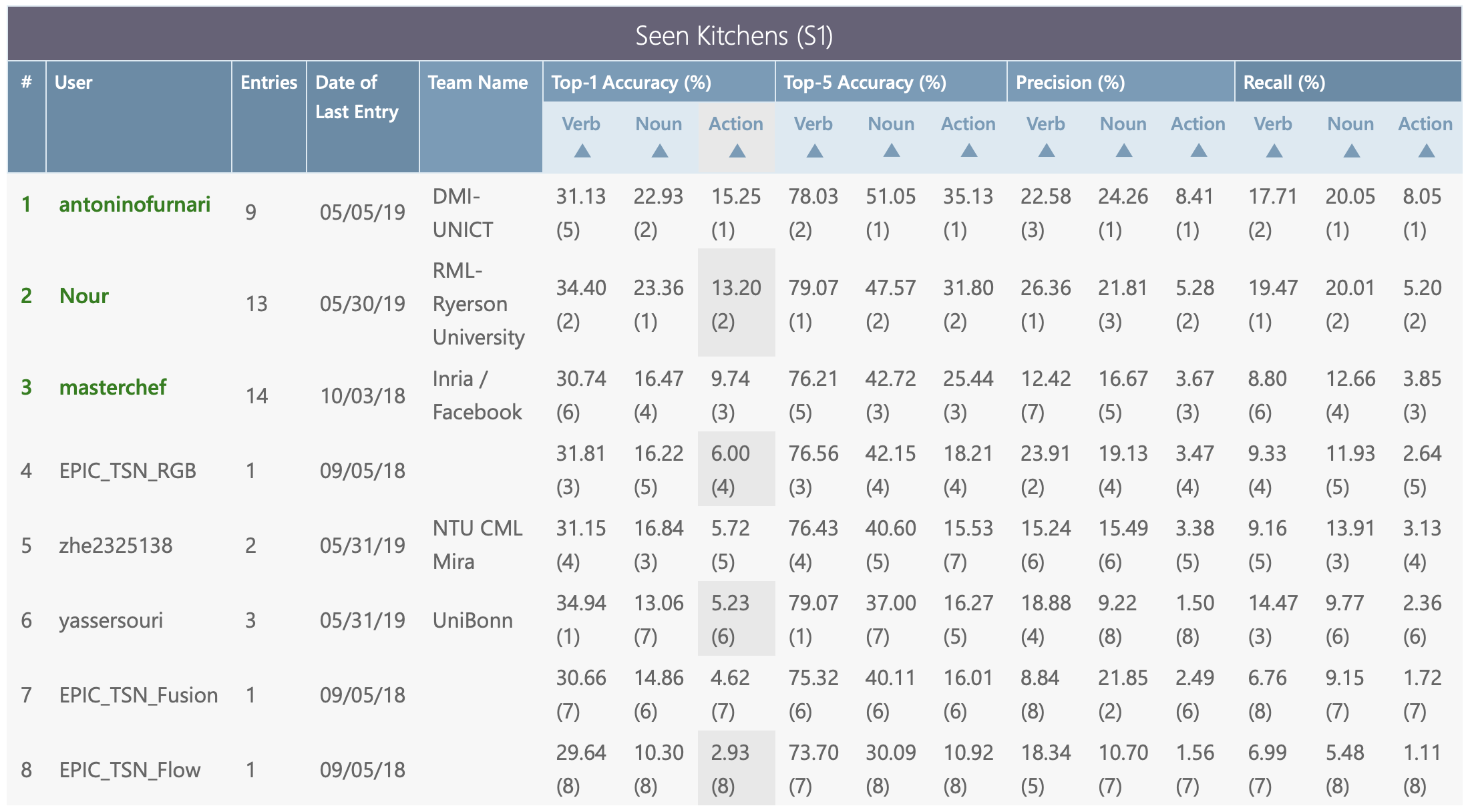}
	\includegraphics[width=\linewidth]{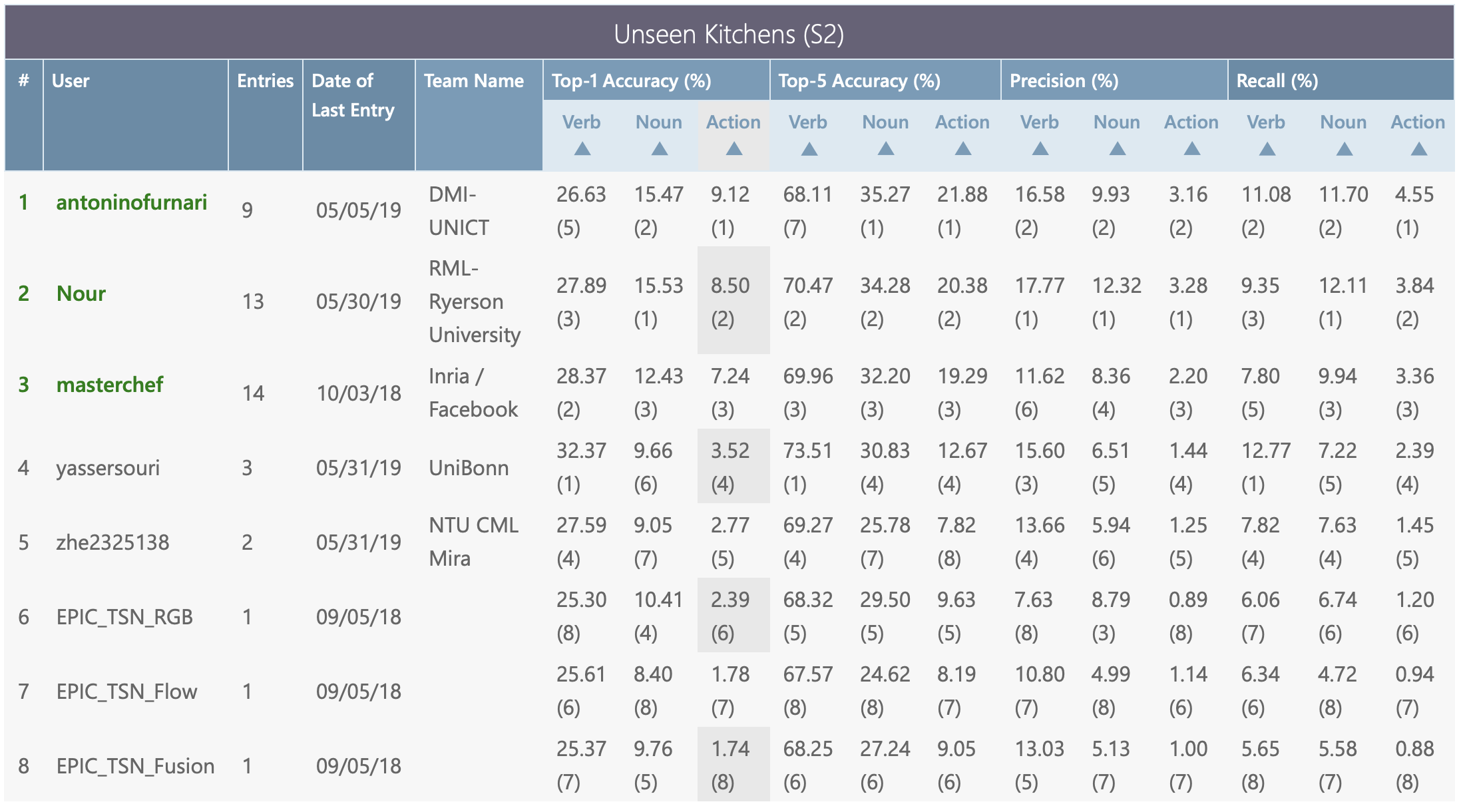}
	\caption{Screenshots of the EPIC-Kitchens Egocentric Action Anticipation Challenge Leaderboards at the end of the competition, acquired from \url{https://epic-kitchens.github.io/} on the $1^{st}$ of August, 2019. The team name of the proposed method is ``DMI-UNICT''.}
	\label{fig:leaderboards}
\end{figure*}

The proposed RULSTM approach has been used to participate in the EPIC-Kitchens egocentric action anticipation competition. Specifically, we considered an ensemble model including features extracted using a BNInception and a ResNet-50 CNN trained for action recognition. \figurename~\ref{fig:leaderboards} reports a screenshot of the EPIC-Kitchens egocentric action anticipation challenge at the end of the competition. The screenshot has been acquired from \url{https://epic-kitchens.github.io/} on the $1^{st}$ of August, 2019. Note that our submission (team name ``DMI-UNICT'') is ranked first on both \textbf{S1} and \textbf{S2}.

\section{Additional Qualitative Examples}
\label{sec:qualitative}
\begin{figure*}
	\includegraphics[width=\linewidth,clip=true,trim=0 0 0 140]{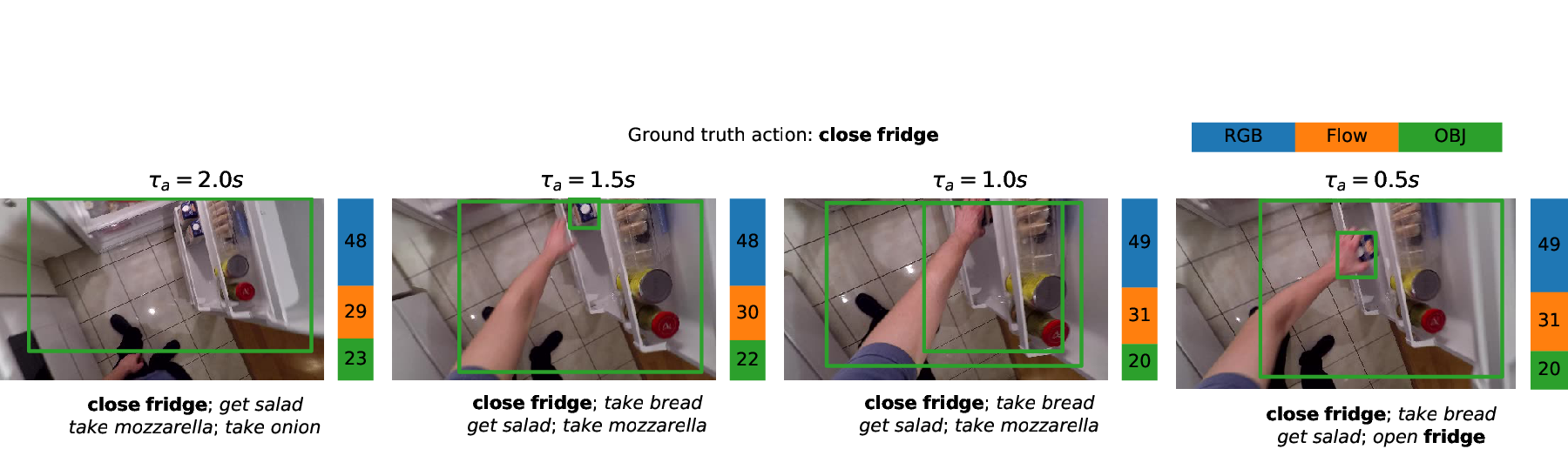}\\
	\includegraphics[width=\linewidth,clip=true,trim=0 0 0 110]{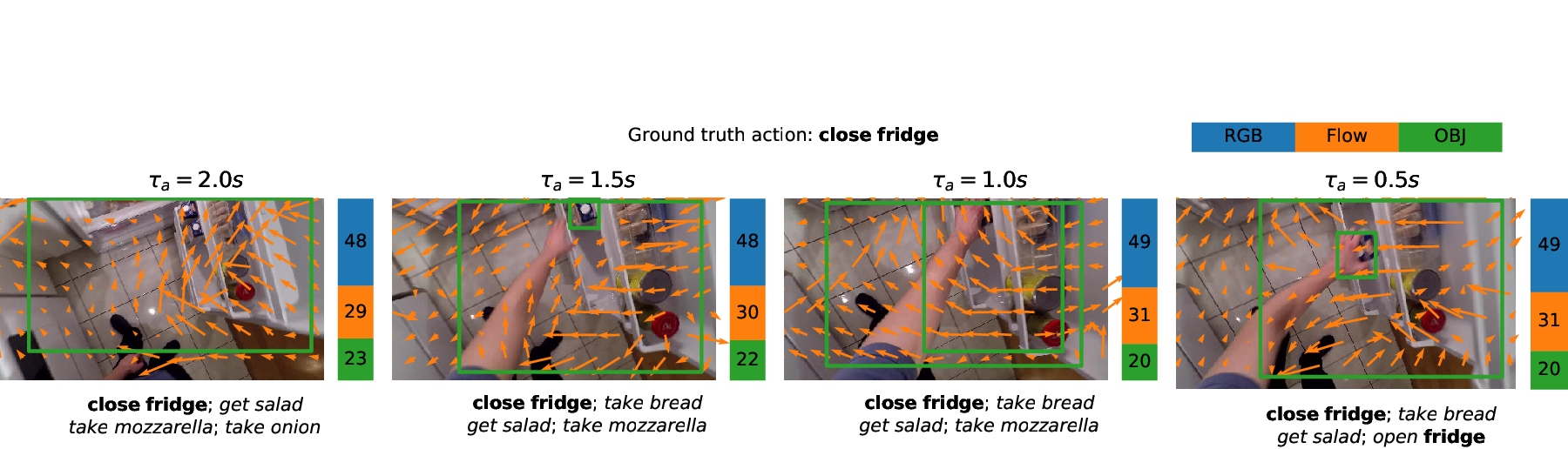}
	
	\includegraphics[width=\linewidth,clip=true,trim=0 0 0 120]{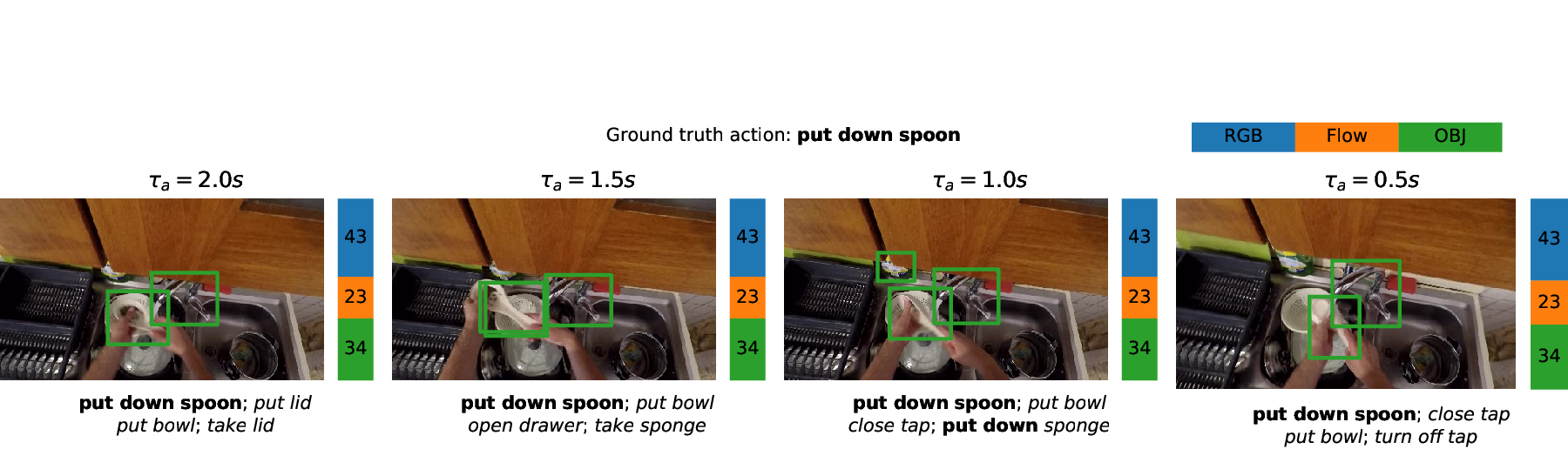}\\
	\includegraphics[width=\linewidth,clip=true,trim=0 0 0 110]{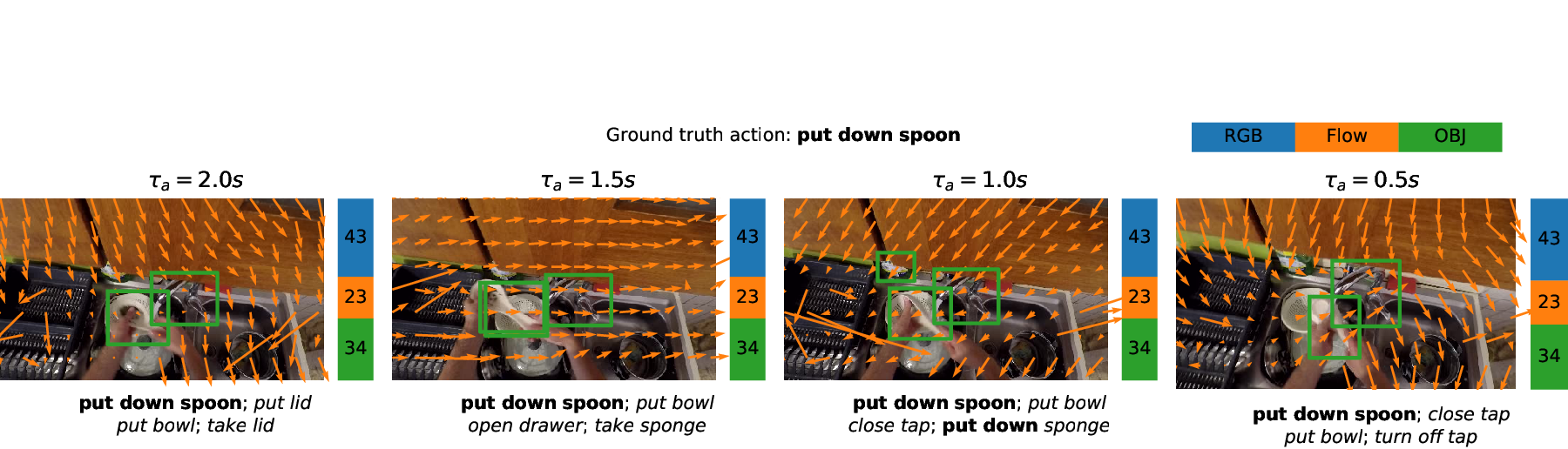}
	
	\includegraphics[width=\linewidth,clip=true,trim=0 0 0 120]{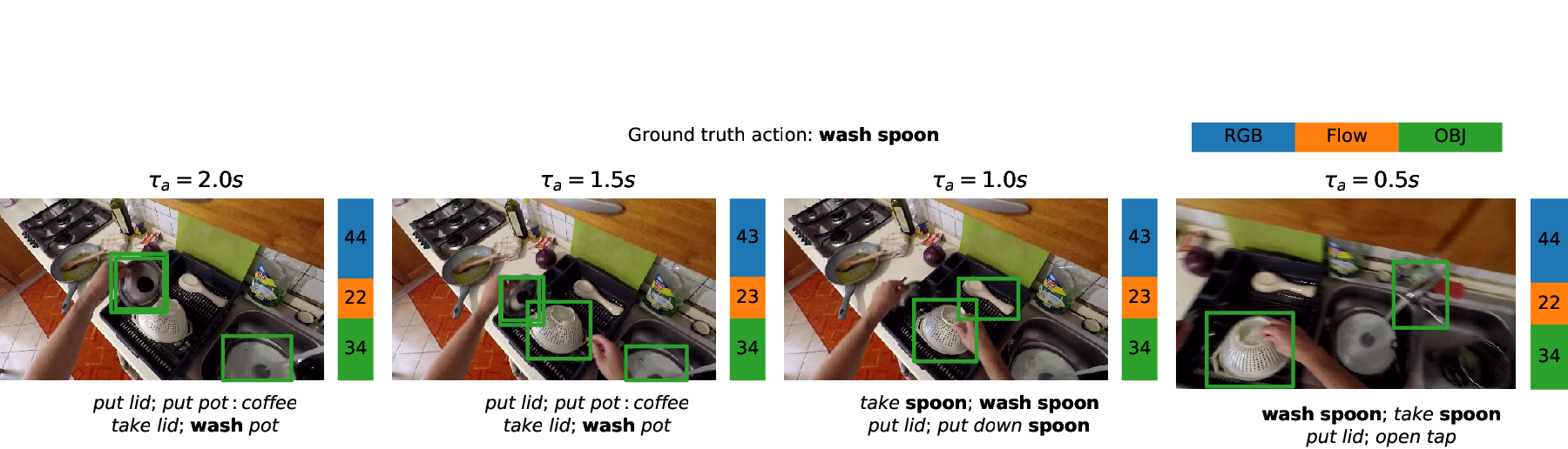}\\
	\includegraphics[width=\linewidth,clip=true,trim=0 0 0 110]{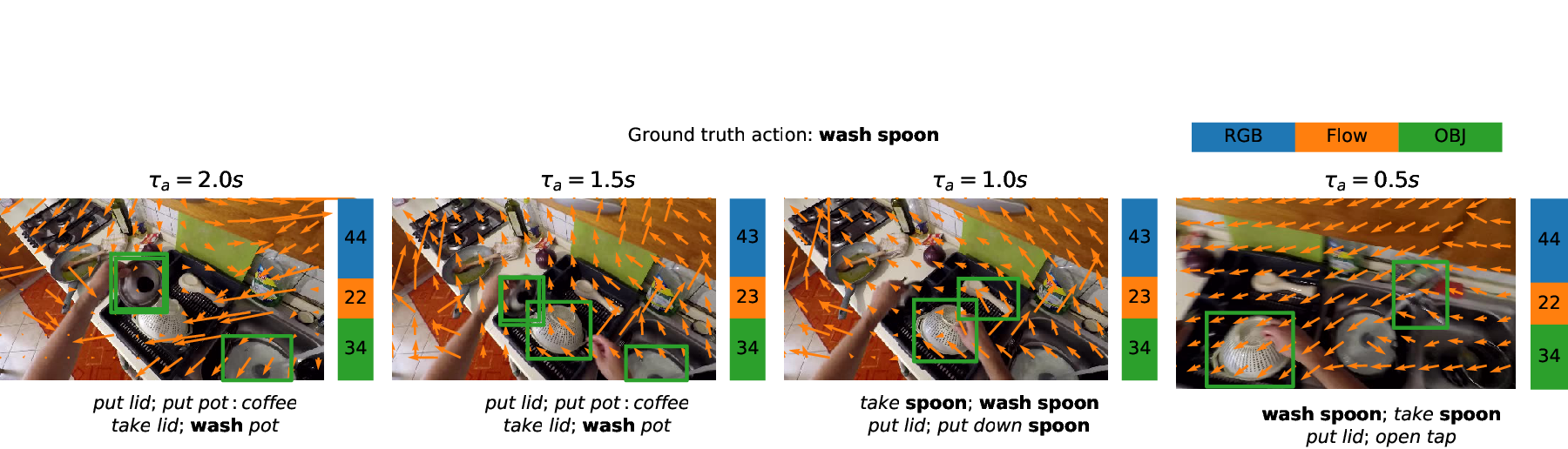}

	\caption{Success action anticipation example qualitative results (best seen on screen).}
	\label{fig:qualitative_success}
\end{figure*}

\figurename{~\ref{fig:qualitative_success}} reports qualitative results of three additional success action anticipation examples. For improved clarity, we report frames with and without optical flows for each example. In the top example, MATT assigns a small weight to the object branch as the contextual appearance features (i.e., RGB) are already enough to reliably anticipate the next actions. In the middle example object detection is fundamental to correctly anticipate ``put down spoon'', as soon as the object is detected. The bottom example shows a complex scene with many objects. The ability to correctly recognize objects is fundamental to anticipate certain actions (i.e., ``wash spoon''). The algorithm can anticipate ``wash'' well in advance. As soon as the spoon is detected ($\tau_a=2s$), ``wash spoon'' is correctly anticipated. Note that, even if the spoon is not correctly detected at time $\tau_a=0.5s$, ``wash spoon'' is still correctly anticipated.

\begin{figure*}
	\includegraphics[width=\linewidth,clip=true,trim=0 0 0 140]{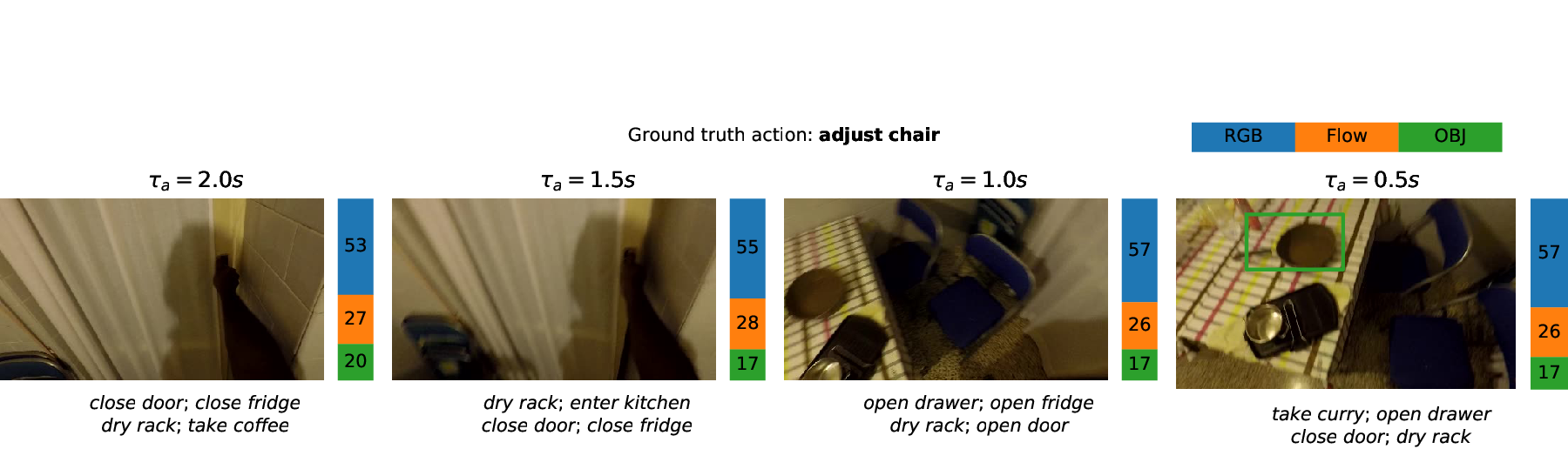}\\
	\includegraphics[width=\linewidth,clip=true,trim=0 0 0 110]{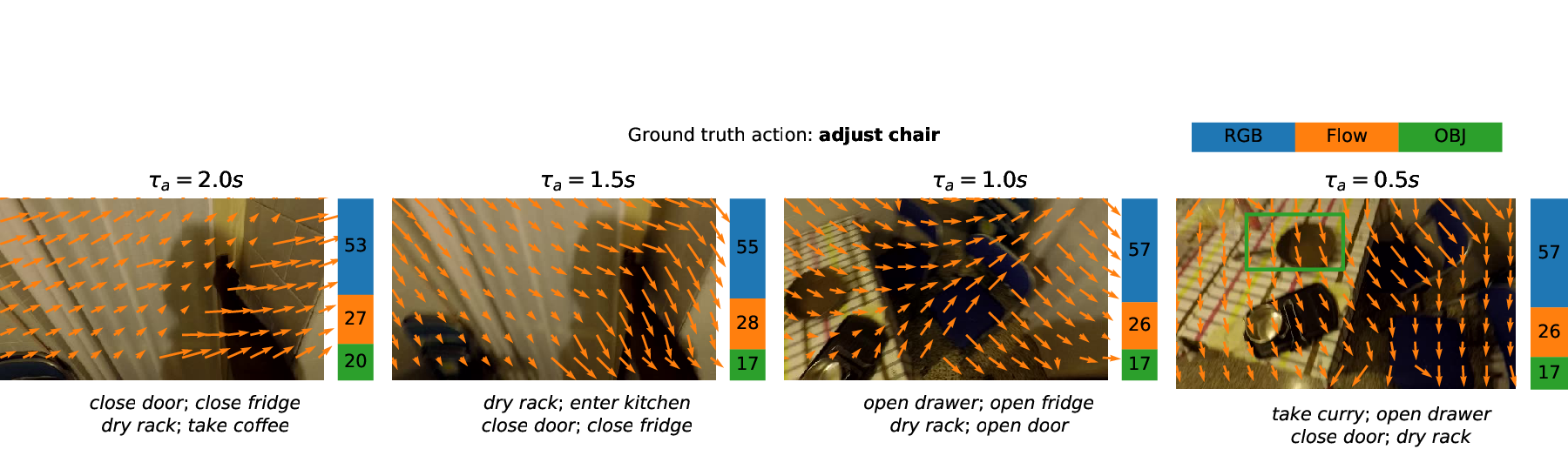}
	
	\includegraphics[width=\linewidth,clip=true,trim=0 0 0 120]{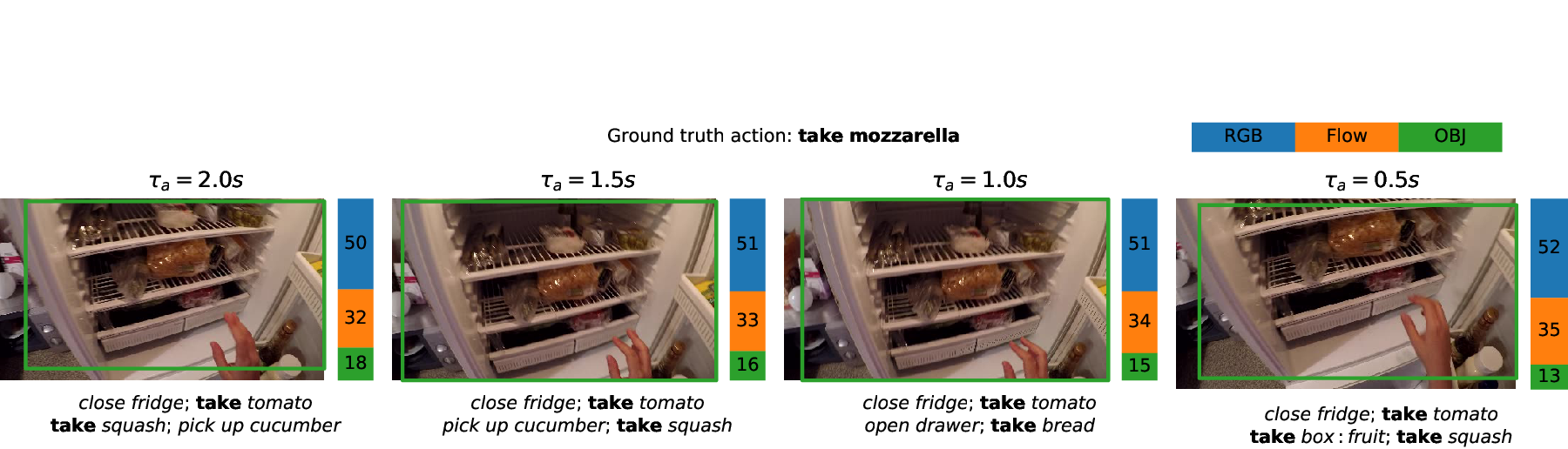}\\
	\includegraphics[width=\linewidth,clip=true,trim=0 0 0 110]{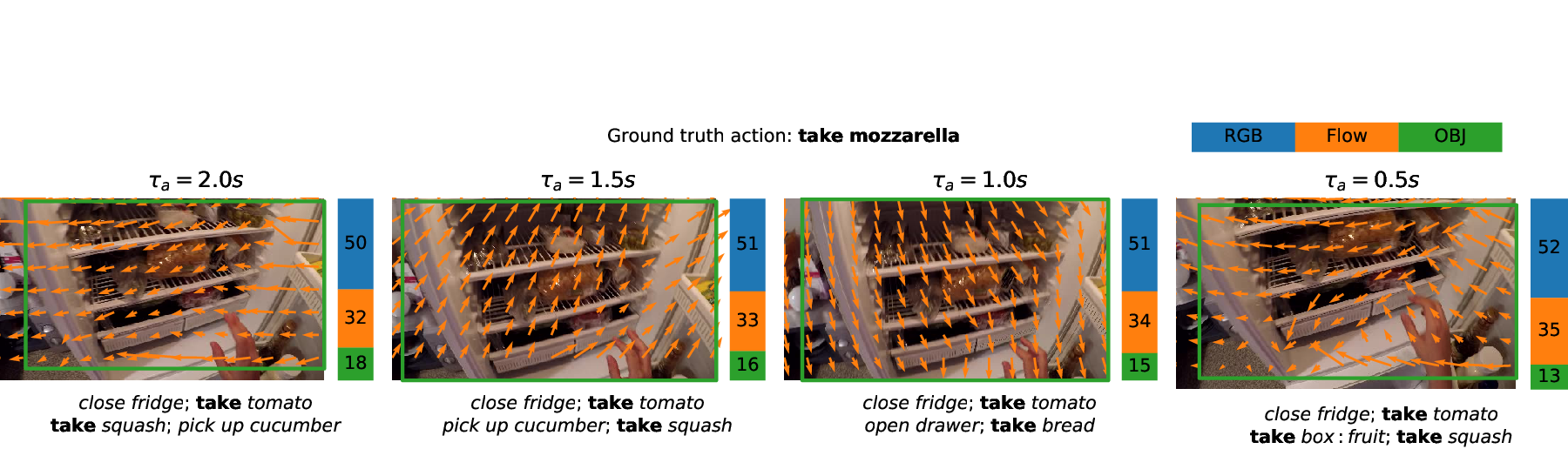}
	
	\includegraphics[width=\linewidth,clip=true,trim=0 0 0 120]{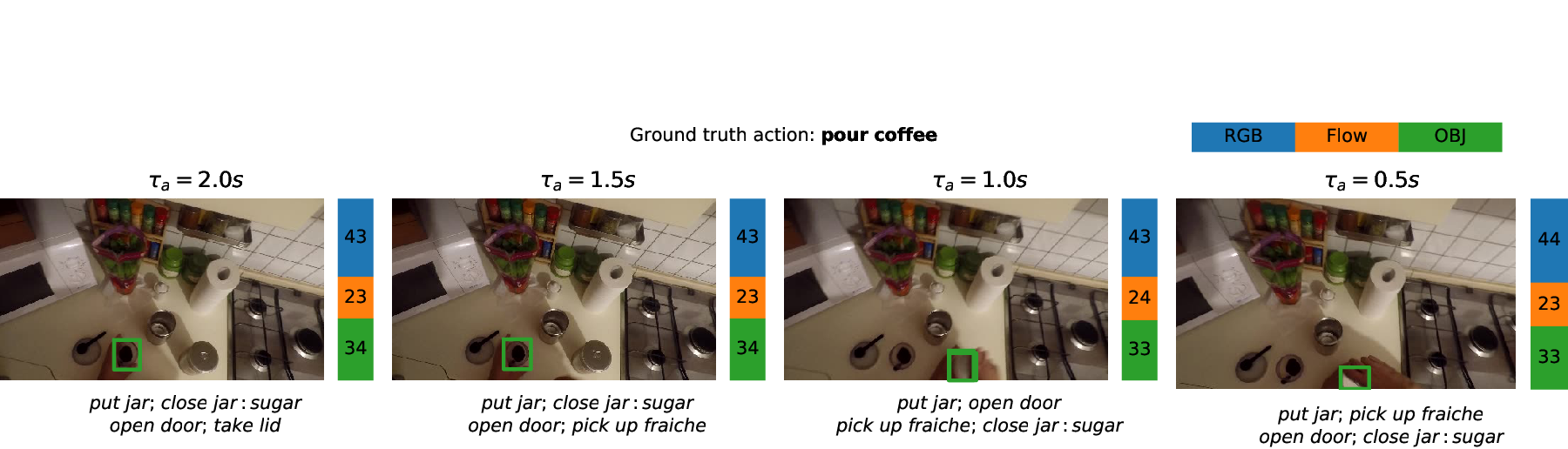}\\
	\includegraphics[width=\linewidth,clip=true,trim=0 0 0 110]{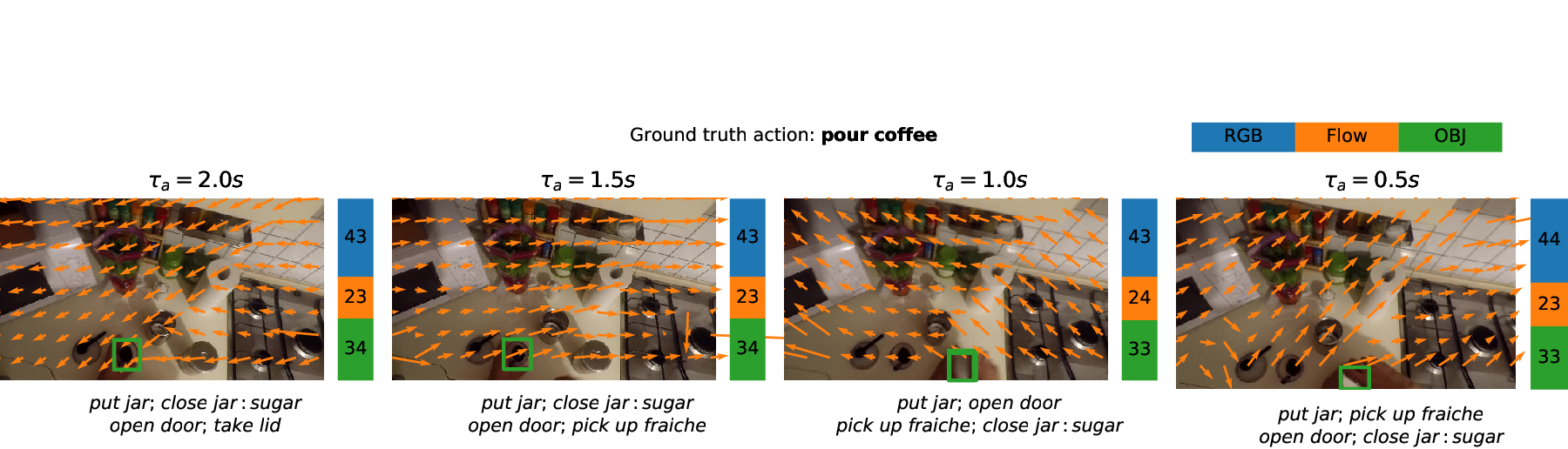}

	\caption{Failure action anticipation example qualitative results (best seen on screen).}
	\label{fig:qualitative_failure}
\end{figure*}

\figurename~\ref{fig:qualitative_failure} reports three failure examples. In the top example, the model fails to predict ``adjust chair'', probably due to the inability of the object detector to identify the chair. Note that, when the object ``pan'' on the table is detected, ``take curry'' is wrongly anticipated. In the middle example, the algorithm successfully detects the fridge and tries to anticipate ``close fridge'' and some actions involving the ``take'' action, with wrong objects. This is probably due to the inability of the detector to detect ``mozzarella'', which is not yet appearing in the scene. In the bottom example, the method tries to anticipate actions involving ``jar'', as soon as ``jar'' is detected. This misleads the algorithm as the correct action is ``pour coffee''.

The reader is referred to the videos in the supplementary material for additional success and failure qualitative examples. The supplementary material also reports qualitative examples of the proposed method when applied to the problem of early action recognition.

{\small
	\bibliographystyle{ieee}
	\bibliography{egbib}
}

\end{document}